\RequirePackage{fix-cm}
\documentclass[twocolumn]{svjour3}          
\smartqed  
\usepackage{graphicx}
\usepackage{latexsym}
\usepackage{times}
\usepackage{epsfig}
\usepackage{graphicx}
\usepackage{amsmath}
\usepackage{amssymb}

\usepackage{graphicx}
\usepackage{amsmath}
\usepackage{amssymb}
\usepackage{booktabs}

\usepackage{subfig}
\usepackage{algorithm}
\usepackage{algorithmic}
\usepackage{multirow}
\usepackage[table,xcdraw]{xcolor}
\usepackage{booktabs}
\usepackage{amsmath}
\usepackage{bbding}
\usepackage{enumitem}

\usepackage[pagebackref=true,breaklinks=true,letterpaper=true,colorlinks,bookmarks=false]{hyperref}

\usepackage{bbding}
\usepackage{tikz}
\usepackage{comment}
\usepackage{amsmath,amssymb} 
\usepackage{color}
\usepackage{graphicx}
\usepackage{comment}
\usepackage{amsmath,amssymb} 
\usepackage{color}
\usepackage{amsmath}
\usepackage{amsmath,amssymb}
\usepackage{multirow}
\usepackage{array}
\usepackage{wrapfig}

\usepackage[T1]{fontenc}
\usepackage[utf8]{inputenc}
\usepackage{babel}
\usepackage[font=small,labelfont=bf]{caption}

\usepackage{makecell}
\usepackage{enumitem}
\usepackage{color}
\usepackage{url}            
\emergencystretch=\maxdimen
\begin{document}

\title{Robust Sequential DeepFake Detection}


\author{Rui Shao         \and
        Tianxing Wu \and
        Ziwei Liu
}


\institute{Rui Shao\at
	School of Computer Science and Technology, Harbin Institute of Technology (Shenzhen)  \\
     shaorui@hit.edu.cn   
           \and
           Tianxing Wu\at 
           S-Lab, Nanyang Technological University \\
         twu012@ntu.edu.sg
         \and
        Ziwei Liu \at 
           S-Lab, Nanyang Technological University \\
         ziwei.liu@ntu.edu.sg \\
         \\
         Ziwei Liu is the corresponding author.
}

\date{Received: date / Accepted: date}

\maketitle

\begin{abstract}
Since photorealistic faces can be readily generated by facial manipulation technologies nowadays, potential malicious abuse of these technologies has drawn great concerns. Numerous deepfake detection methods are thus proposed. However, existing methods only focus on detecting \textit{one-step} facial manipulation. As the emergence of easy-accessible facial editing applications, people can easily manipulate facial components using \textit{multi-step} operations in a sequential manner. This new threat requires us to detect a sequence of facial manipulations, which is vital for both detecting deepfake media and recovering original faces afterwards. Motivated by this observation, we emphasize the need and propose a novel research problem called Detecting Sequential DeepFake Manipulation (\textbf{Seq-DeepFake}). Unlike the existing deepfake detection task only demanding a binary label prediction, detecting Seq-DeepFake manipulation requires correctly predicting a sequential vector of facial manipulation operations. To support a large-scale investigation, we construct the first Seq-DeepFake dataset, where face images are manipulated sequentially with corresponding annotations of sequential facial manipulation vectors. Based on this new dataset, we cast detecting Seq-DeepFake manipulation as a specific image-to-sequence (\textit{e.g.,} image captioning) task and propose a concise yet effective Seq-DeepFake Transformer (\textbf{SeqFakeFormer}). \textcolor{black}{To better reflect real-world deepfake data distributions, we further apply various perturbations on the original Seq-DeepFake dataset and construct the more challenging Sequential DeepFake dataset with perturbations (\textbf{Seq-DeepFake-P}). To exploit deeper correlation between images and sequences when facing Seq-DeepFake-P, a dedicated Seq-DeepFake Transformer with Image-Sequence Reasoning (\textbf{SeqFakeFormer++}) is devised, which builds stronger correspondence between image-sequence pairs for more robust Seq-DeepFake detection. Moreover, we build a comprehensive benchmark and set up rigorous evaluation protocols and metrics for this new research problem. Extensive quantitative and qualitative experiments demonstrate the effectiveness of SeqFakeFormer and SeqFakeFormer++.} Several valuable observations are also revealed to facilitate future research in broader deepfake detection problems. The code has been released at \href{https://github.com/rshaojimmy/SeqDeepFake/} {https://github.com/rshaojimmy/SeqDeepFake/}.
\end{abstract}

\section{Introduction}

\begin{figure*}[t] 
	\begin{center}
		\includegraphics[width=0.83\linewidth]{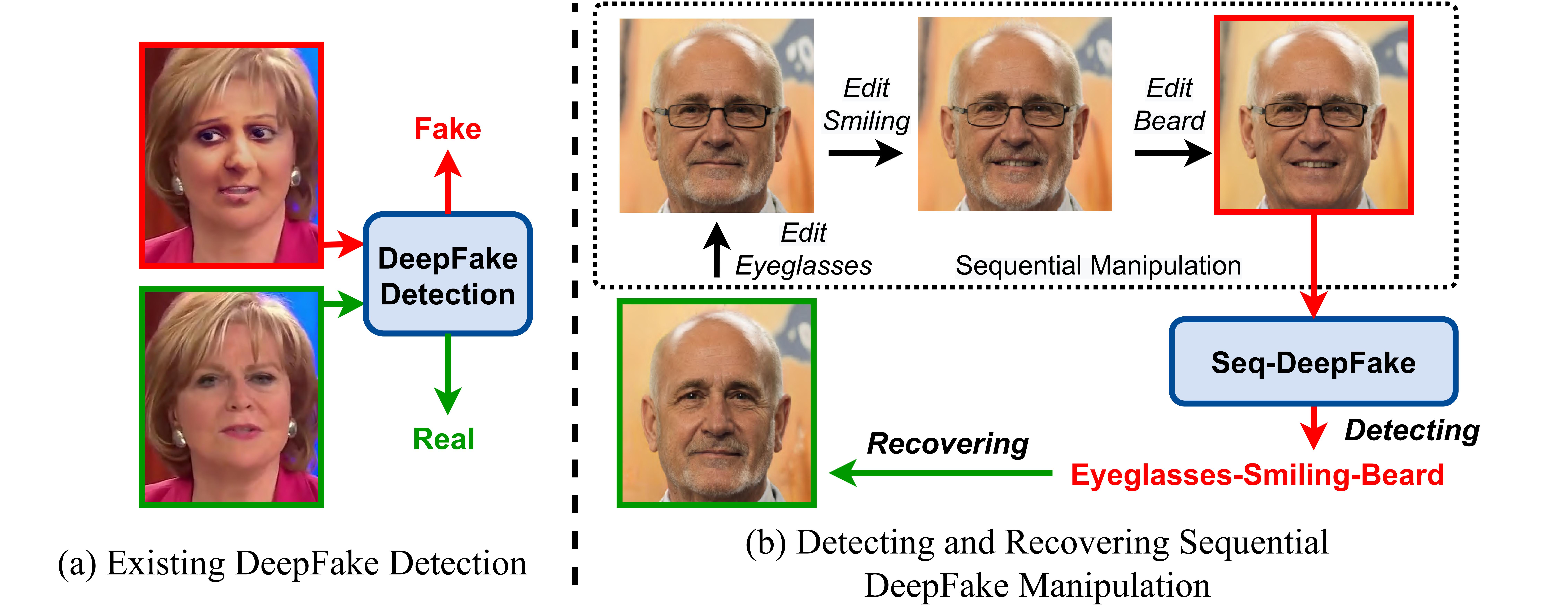}
	\end{center}
	\caption{Comparison between (a) existing deepfake detection and (b) proposed detecting and recovering sequential deepfake manipulation.}
	\label{fig:deepfakecmp}
\end{figure*}

In recent years, hyper-realistic face images can be generated by deep generative models~\cite{li2024optimus,chen2024lion,shen2024mome} which are visually extremely indistinguishable from real images. Meanwhile, the significant improvement for image synthesis brings security concerns on potential malicious abuse of these techniques that produce misinformation and fabrication, which is known as \textit{deepfake}~\cite{shao2023deepfake,shao2023detecting,shao2024detecting}. To address this security issue, various deepfake detection methods have been proposed to detect such forged faces. As illustrated in Fig.~\ref{fig:deepfakecmp} (a), given the manipulated face image generated by face swap algorithm~\cite{rossler2019faceforensics++} and the original face image, the existing deepfake detection task requires the model to predict the correct binary labels (Real/Fake).

With the increasing popularity of easy-accessible facial editing applications, such as YouCam Makeup\footnote{\tiny{https://apps.apple.com/us/app/youcam-makeup-selfie-editor/id863844475}}, FaceTune2\footnote{\tiny{https://apps.apple.com/us/app/facetune2-editor-by-lightricks/id1149994032}}, and YouCam Perfect\footnote{\tiny{https://apps.apple.com/us/app/youcam-perfect-photo-editor/id768469908}}, it is convenient for people to edit face images in daily life. Compared to existing deepfake techniques mainly carrying out \textit{one-step} facial manipulation~\cite{rossler2019faceforensics++,he2021forgerynet}, we can now easily manipulate face images using \textit{multi-step} operations in a \textit{sequential} manner. As shown in Fig.~\ref{fig:deepfakecmp} (b), the original image can be manipulated by adding eyeglasses, making a bigger smile and removing beard sequentially. This expands the scope of existing deepfake problem by adding sequential manipulation information and poses a new challenge for current \textit{one-step} deepfake detection methods. This observation motivates us to introduce a new research problem --- Detecting Sequential DeepFake Manipulation (\textbf{Seq-DeepFake}). We summarize several key differences between detecting Seq-DeepFake and the existing deepfake detection: \textbf{1)} Rather than only predicting binary labels (Real/Fake), detecting Seq-DeepFake aims to detect sequences of facial manipulations with diverse sequence lengths. For example, the model is required to predict a 3-length sequence as `Eyeglasses-Smiling-Beard' for the manipulated image as shown in Fig.~\ref{fig:deepfakecmp} (b) and \textbf{2)} As illustrated in Fig.~\ref{fig:deepfakecmp} (b), beyond pure forgery detection, we can further \textit{recover} the original faces (refer to Sec.~\ref{sec:facerecovery} in Experiments) based on the detected sequences of facial manipulation in Seq-DeepFake. This greatly enriches the benefits of detecting Seq-DeepFake manipulation.

\begin{figure}[t] 
	\begin{center}
		\includegraphics[width=1\linewidth]{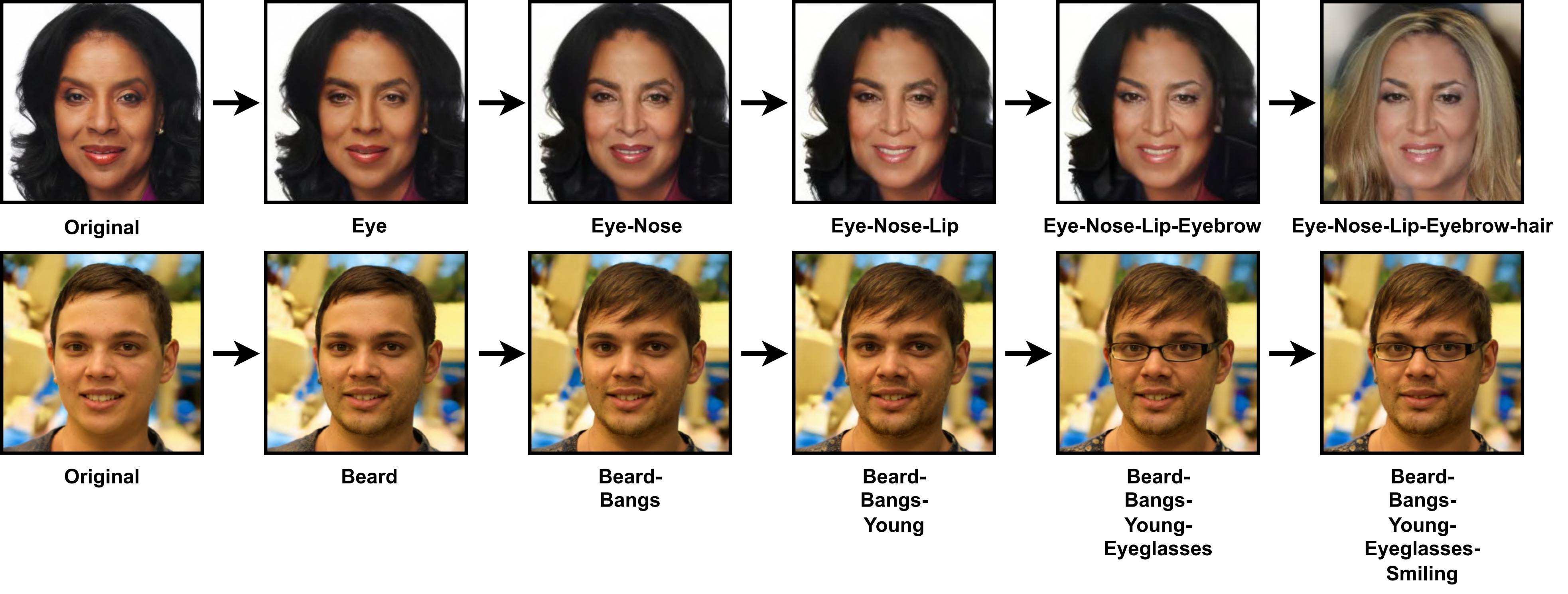}
	\end{center}
	\vspace*{-3mm}
	\caption{Illustration of sequential facial manipulation. Two types of facial manipulation approaches are considered, \textit{i.e.,} facial components manipulation~\cite{kim2021exploiting} in the first row and facial attributes manipulation~\cite{jiang2021talk} in the second row.}
	\label{fig:DatasetIntro}
	\vspace{-15pt}
\end{figure}

To facilitate the study of detecting Seq-DeepFake, this paper contributes the first Seq-DeepFake dataset. Fig.~\ref{fig:DatasetIntro} shows some samples in Seq-DeepFake dataset. From Fig.~\ref{fig:DatasetIntro}, it can be seen that one face image can be sequentially manipulated with different number of steps (from minimum 1 step to maximum 5 steps), which leads to facial manipulation sequences with diverse lengths. It is extremely difficult to distinguish the original and manipulated face images, and even harder to figure out the exact manipulation sequences. To make our study more comprehensive, we consider two different facial manipulation techniques, \textit{i.e.,} facial components manipulation~\cite{kim2021exploiting} and facial attributes manipulation~\cite{jiang2021talk}, which are displayed in the first and second row, respectively in Fig.~\ref{fig:DatasetIntro}.

Most of current facial manipulation applications are built based on Generative Adversarial Network (GAN). It is well known that the semantic latent space learned by GAN is difficult to be perfectly disentangled~\cite{shen2020interpreting,lee2020high}. We argue that this defect is likely to leave some spatial as well as sequential manipulation traces unveiling sequential facial manipulations. Based on this observation, to 
detect such two types of manipulations traces, we cast detecting Seq-DeepFake as a specific image-to-sequence (\textit{e.g.,} image captioning) task and thus propose a concise yet effective Seq-DeepFake Transformer (\textbf{SeqFakeFormer}). Two key components
are devised in SeqFakeFormer: \textbf{Spatial Relation Extraction} via Image Encoder and \textbf{Sequential Relation Modeling with Spatially Enhanced Cross-Attention} via Sequence Decoder. Given a manipulated image, to adaptively capture subtle spatial manipulation regions, SeqFakeFormer feeds the image into a deep convolutional neural network (CNN) to learn its feature maps. Then we extract the relation of spatial manipulations captured in feature maps using the self-attention modules of Image Encoder, obtaining features of spatial relation, \textit{i.e.,} spatial manipulation traces. After that, Sequence Decoder of SeqFakeFormer models the sequential relation of extracted spatial relation features via cross-attention modules in an autoregressive mechanism. This captures sequential manipulation traces, and thereby detects the facial manipulation sequences. To enable more effective cross-attention given limited annotations of facial manipulation sequences, SeqFakeFormer further integrates a Spatially Enhanced Cross-Attention (SECA) module in the decoder. This module enriches the spatial information of manipulation sequences by learning a spatial weight map. After fusing the spatial weight map with the cross-attention map, a spatially enhanced cross-attention can be achieved.

\textcolor{black}{To better imitate deepfake data distributions in real-world scenarios, we further apply 6 types of perturbations, such as color distortion, multiplicative noise, random compression and blur, to the original Seq-DeepFake data at 3 intensity levels. This constructs a more challenging version of Seq-DeepFake dataset with perturbations (\textbf{Seq-DeepFake-P}). To deal with such a more difficult case, we further propose a more dedicated model dubbed Seq-DeepFake Transformer with Image-Sequence Reasoning (\textbf{SeqFakeFormer++}). Apart from the extraction of image-sequence correlation via cross-attention in SeqFakeFormer, SeqFakeFormer++ further performs deeper correlation reasoning between images and sequences via \textbf{Image-Sequence Contrastive Learning (ISC)} and \textbf{Image-Sequence Matching (ISM)}. The former reasoning module aligns the features of spatial relation and features of sequential relation by pulling those of matched image-sequence pairs close while pushing those of unmatched pairs apart. Meanwhile, the latter one achieves more fine-grained feature alignment between images and sequences by predicting whether an image-sequence pair is matched or unmatched based on aggregated multi-modal features. Aided by these two reasoning modules, more robust cross-modal correlation between images and sequences could be built and thus SeqFakeFormer++ is more able to unveil sequence manipulation traces under severe post-processing perturbations.}

Main contributions of our paper can be summarized as:

\begin{itemize}[leftmargin=*]
\item We introduce a new research problem named Detecting Sequential DeepFake Manipulation (\textbf{Seq-DeepFake}), with the objective of detecting sequences of facial manipulations, which expands the scope and poses a new challenge for deepfake detection.

\item We contribute the first large-scale Seq-DeepFake dataset generating sequential manipulated face images using two different facial manipulation techniques. Corresponding annotations of manipulation sequences are provided. \textcolor{black}{To mimic various noisy scenarios in real-world, we further apply a series of perturbations on manipulated face images and construct the more challenging Seq-DeepFake dataset with perturbations (\textbf{Seq-DeepFake-P}).}

\item We propose a powerful Seq-DeepFake Transformer (\textbf{SeqFakeFormer}). \textcolor{black}{Aiming at better detecting sequence of deepfake manipulation under perturbations, a Seq-DeepFake Transformer with Image-Sequence Reasoning (\textbf{SeqFakeFormer++}) is devised, providing deeper cross-modal reasoning for more robust detection under perturbations.} A comprehensive benchmark is built and rigorous evaluation protocols and metrics are designed for this novel research problem. \textcolor{black}{Extensive quantitative and qualitative experiments demonstrate the superiority of SeqFakeFormer and SeqFakeFormer++.}
\end{itemize}

\textcolor{black}{This paper is a substantial extension of our previous ECCV 2022 work~\cite{shao2022detecting}. We have made three major improvements in this journal version: \textbf{1)} We construct a more realistic and challenging Seq-DeepFake dataset with perturbations (\textbf{Seq-DeepFake-P}). It is designed to evaluate the robustness of Seq-DeepFake detection to various perturbations in real-world; \textbf{2)} To promote such robustness, a more powerful Seq-DeepFake Transformer with Image-Sequence Reasoning (\textbf{SeqFakeFormer++}) is developed. Two more reasoning modules, Image-Sequence Contrastive Learning via Sequence Encoder and Image-Sequence Matching via Image-attended Sequence Encoder, are integrated and complemented along with Sequence Decoder; \textbf{3)} More visualizations corresponding to Seq-DeepFake and Seq-DeepFake-P datasets are provided. More detailed and comprehensive quantitative and qualitative experiments regarding SeqFakeFormer and SeqFakeFormer++ are performed in this journal version.}

\section{Related Work}
\subsection{DeepFake Detection} 
The issue of facial information security~\cite{2018TIFSdynamictext,Shao_2019_CVPR,Shao_2020_AAAI,shao2022open,shao2020open,shao2017deep,shao2022federated} is receiving increasing attention. Current deepfake detection methodscan be roughly categorized into spatial-based and frequency-based deepfake detection. The majority of spatial-based deepfake detection methods focus on capturing visual cues from spatial domain. Face X-ray~\cite{li2020face} is proposed to detect the blending boundary left in the face forgery process as visual cues for real/fake detection. A multi-attentional deepfake detection network is proposed in~\cite{zhao2021multi} to integrate low-level textural features and high-level semantic features. Zhu \textit{et al.}~\cite{zhu2021face} introduce 3D decomposition into forgery detection and propose a two-stream network to fuse decomposed features for detection. Pair-wise self-consistency learning (PCL)~\cite{zhao2021learning} is introduced to detect inconsistency of source features within the manipulated images. Inconsistencies in semantically high-level mouth movements are captured in ~\cite{haliassos2021lips} by fine-tuning a temporal network pretrained on lipreading. On the other hand, some methods pay attention to the frequency domain for detecting spectrum artifacts. There exist distinct spectrum distributions and characteristics between real and fake images in the high-frequency part of Discrete Fourier Transform (DFT)~\cite{durall2019unmasking,dzanic2020fourier}. Qian \textit{et al.}~\cite{qian2020thinking} propose a F$^{3}$-Net to learn local frequency statistics based on Discrete Cosine Transform (DCT) to mine forgery. Liu \textit{et al.}~\cite{liu2021spatial} present a Spatial-Phase Shallow Learning method to fuse spatial image and phase spectrum for the up-sampling artifacts detection. A two-stream model is devised in~\cite{luo2021generalizing} to model the correlation between extracted high-frequency features and regular RGB features to learn generalizable features. A frequency-aware discriminative feature learning framework~\cite{li2021frequency} is introduced to integrate metric learning and adaptive frequency features learning for face forgery detection.

So far, several deepfake datasets have been released to public, such as FaceForensics++~\cite{rossler2019faceforensics++}, Celeb-DF~\cite{li2020celeb}, DeepFake Detection Challenge (DFDC)~\cite{dolhansky2019deepfake}, and DeeperForensics-1.0 (DF1.0)~\cite{jiang2020deeperforensics}. However, only binary labels are provided in most of existing deepfake datasets, and thus most of the above works are trained to carry out binary classification, which results in performance saturation and poor generalization.
\subsection{Facial Editing} 
Several methods have been proposed for editing facial components (\textit{i.e.} eye, nose, month). Lee \textit{et al.}~\cite{lee2020maskgan} present a geometry-oriented face manipulation network MaskGAN for diverse and interactive face manipulation guided by semantic masks annotations. A semantic region-adaptive normalization (SEAN)~\cite{zhu2020sean} is proposed to facilitate manipulating face images by encoding images into the per-region style codes conditioned on segmentation masks. StyleMapGAN~\cite{kim2021exploiting} introduces explicit spatial dimensions to the latent space and manipulates facial components by blending the latent spaces between reference and original faces. Moreover, some works target editing specific facial attributes such as age progression~\cite{yang2018learning}, and smile generation~\cite{wang2018every}. Some recent works discover semantically meaningful directions in the latent space of a pretrained GAN so as to carry out facial attributes editing by moving the latent code along these directions~\cite{shen2020interpreting,shen2020interfacegan,zhuang2021enjoy,voynov2020unsupervised,shen2021closed}. InterFaceGAN~\cite{shen2020interpreting,shen2020interfacegan} tries to disentangle attribute representations in the latent space of GANs by searching a hyperplane, of which a normal vector is used as the editing direction. Fine-grained facial attributes editing is achieved by~\cite{jiang2021talk} through searching a curving trajectory with respect to attribute landscapes in the latent space of GANs.
\section{Sequential DeepFake Dataset}
\label{sec:seq-deepfake-dataset}
\begin{figure*}[t] 
	\begin{center}
		\includegraphics[width=0.85\linewidth]{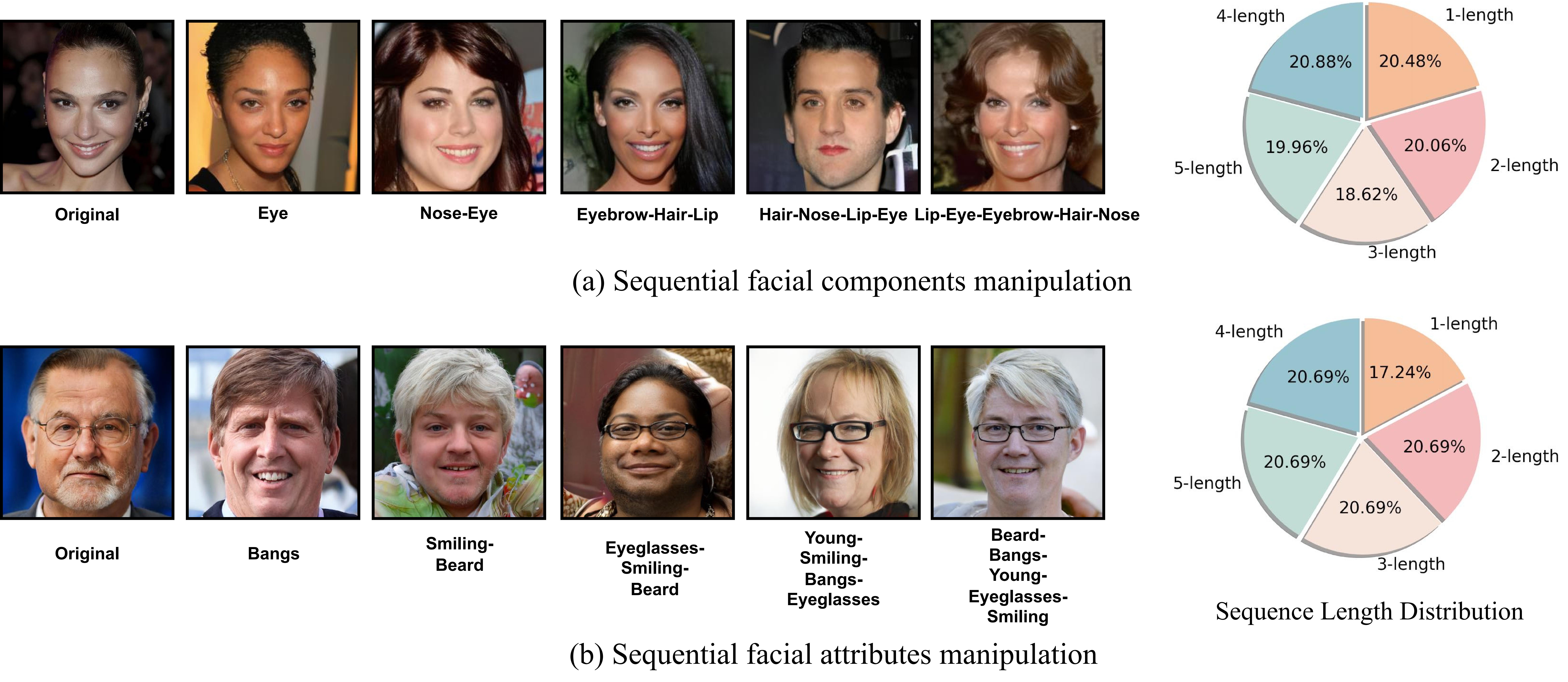}
	\end{center}
	\caption{Illustration of Seq-DeepFake dataset. Samples of Seq-DeepFake are provided with annotations of manipulation sequences. We also show the distribution of sequence length in Seq-DeepFake dataset.}
	\label{fig:DatasetExp}
\end{figure*}
\begin{figure}[t]
	\begin{center}
		\includegraphics[width=1\linewidth]{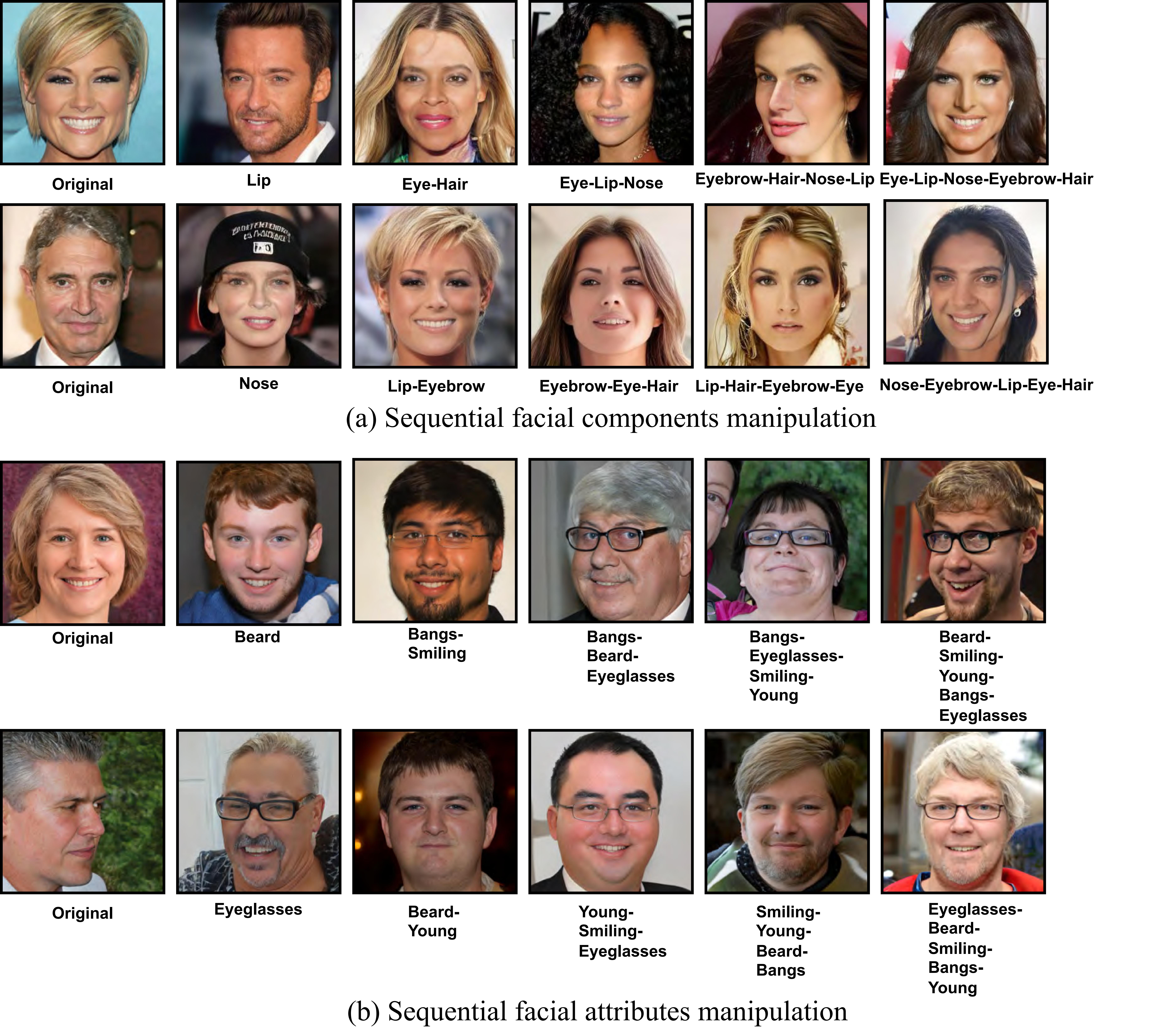}
	\end{center}
	\caption{\textcolor{black}{More samples of Seq-DeepFake dataset. Various sequential facial manipulations are produced with diverse manipulation steps, expressions, ages, and genders.}}
	\label{fig:DatasetExp-appendix}
    \vspace{-13pt}
\end{figure}
To support the novel research problem, we generate a large-scale Sequential DeepFake (Seq-DeepFake) dataset consisting of sequential manipulated face images based on two representative facial manipulation techniques, facial components manipulation~\cite{kim2021exploiting} and facial attributes manipulation~\cite{jiang2021talk}. Unlike most of existing deepfake datasets~\cite{rossler2019faceforensics++,he2021forgerynet} only providing binary labels, the proposed dataset contains annotations of manipulation sequences with diverse sequence lengths. Details of generation pipelines based on the two facial manipulation techniques are as follows.
\subsection{Sequential facial components manipulation}
We adopt the StyleMapGAN proposed in~\cite{kim2021exploiting} for facial components manipulation. Facial components manipulation is carried out based on original images from CelebA-HQ~\cite{liu2015deep,karras2017progressive} and corresponding facial component masks from CelebAMask-HQ~\cite{lee2020maskgan} dataset. Facial components manipulation aims to transplant some facial components of a reference image to an original image with respect to a mask that indicates the components to be manipulated. Specifically, we project the original image and the reference image through the encoder of StyleMapGAN to obtain stylemaps, which are intermediate latent spaces with spatial dimensions. Then, the facial components manipulation is carried out by blending the stylemaps extracted from reference and original faces based on facial component masks. Due to the inevitable appearance of degraded images in the generation process, we adopt the Generated Image Quality Assessment (GIQA) algorithm~\cite{gu2020giqa} to quantitatively evaluate the quality of each generated image and then filter out some low-quality ones based on the pre-defined threshold. Fig.~\ref{fig:DatasetExp} (a) shows some samples with corresponding annotations of sequential facial components manipulation. Through this data generation pipeline, we can finally generate 35,166 manipulated face images annotated with 28 types of manipulation sequences in different lengths (including original). As illustrated in Fig.~\ref{fig:DatasetExp} (a), the proportions of 1-5 different lengths of manipulation sequences are: 20.48\%, 20.06\%, 18.62\%, 20.88\%, 19.96\%.
\subsection{Sequential facial attributes manipulation} Unlike facial components manipulation methods that swap certain local parts from a reference image to an original image, facial attributes manipulation approaches directly change specific attributes on the original face image without any reference images. To take this manipulation type into consideration, we utilize the fine-grained facial editing method proposed by~\cite{jiang2021talk}. This method aims to learn a location-specific semantic field for each editing type on the training set, then edit this attribute of interest on the given face image to a user-defined degree by stepping forward or backward on the learned curve in latent space. Based on this idea, we further generate face images with sequential facial attributes manipulation by performing the editing process in a sequential manner. Specifically, we first sample latent codes from the StyleGAN trained on FFHQ dataset~\cite{karras2019style} to generate original images. Then according to pre-defined attribute sequences, we progressively manipulate each attribute on the original face to another randomly chosen degree using the above method. After generating the final manipulation results, we also perform GIQA algorithm to filter out low-quality samples. Using this pipeline, we generate 49,920 face images with 26 manipulation sequence types, with the length of each sequence ranging from 1 to 5. Since this generation pipeline is more controllable than facial components manipulation, we construct a more balanced dataset, as shown in Fig.~\ref{fig:DatasetExp} (b).
\textcolor{black}{More samples generated by the above two types of manipulation approaches in Seq-DeepFake dataset are displayed in Fig.~\ref{fig:DatasetExp-appendix}. It can be seen from Fig.~\ref{fig:DatasetExp} and~\ref{fig:DatasetExp-appendix} that various sequential facial manipulations are produced with diverse manipulation steps, expressions, ages, and genders.}
\textcolor{black}{\section{Sequential DeepFake Dataset with Perturbations}}
\begin{figure}[t] 
	\begin{center}
		\includegraphics[ width=1\linewidth]{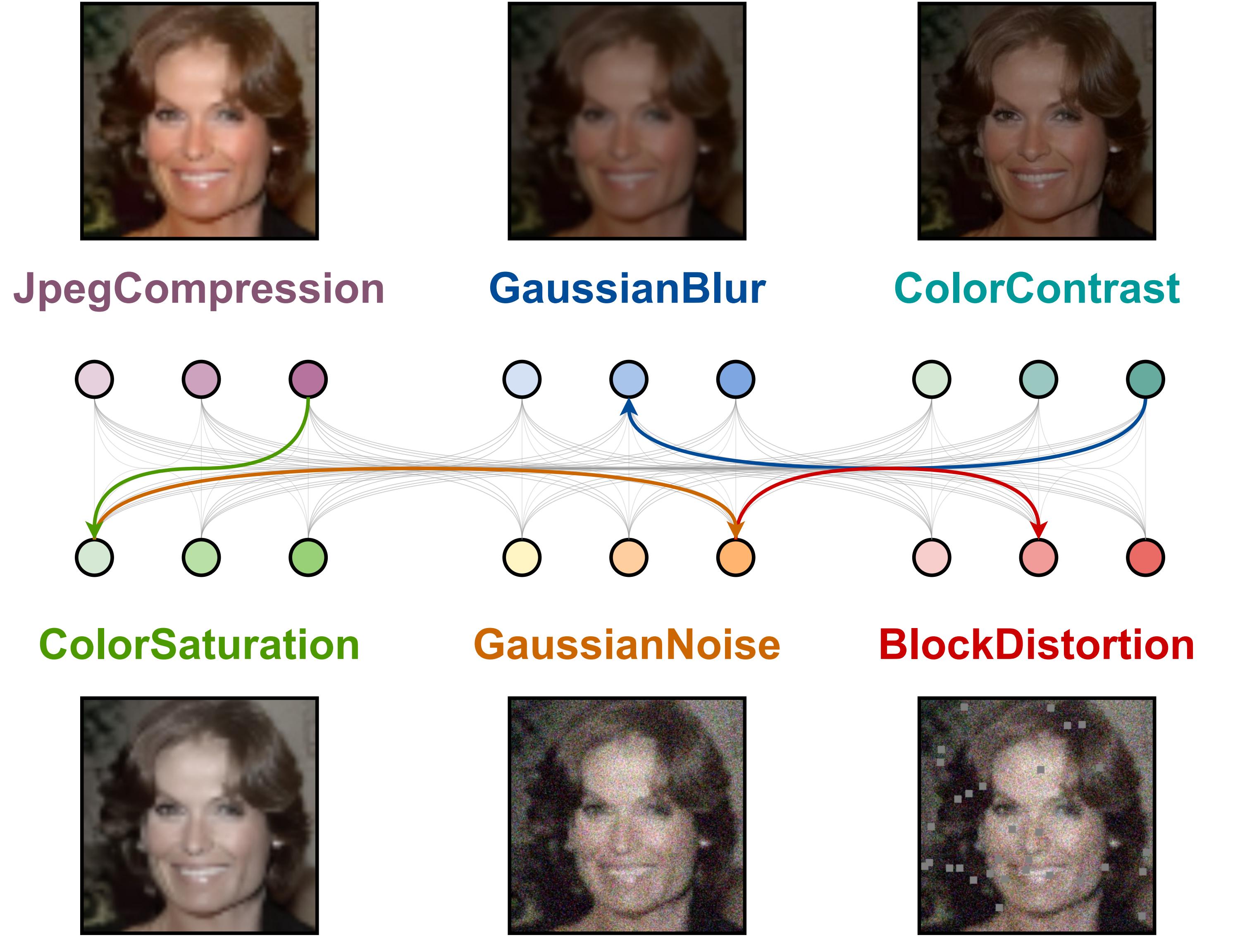}
	\end{center}
	\caption{\textcolor{black}{Illustration of mixing process of perturbations in Seq-DeepFake-P dataset. Different perturbation types and intensity levels are marked in different colors. Arrows represent the mixture order, \textit{e.g.,} the image on the top-middle is first added Color Contrast Change then followed by Gaussian Blur.}}
	\label{fig:dataset_perturb}
	\vspace{-10pt}
\end{figure}
\textcolor{black}{In real-world scenarios, face images are likely to be affected by various post-processing like compression and noise. These post-processing operations degrade the visual quality of face images and tend to cover or alter the manipulation traces. This brings greater challenges for detecting Seq-DeepFake. To better reflect real-world deepfake data distributions under this more realistic scenario, we further impose a series of perturbations on the original Seq-DeepFake dataset and construct a more challenging Sequential DeepFake dataset with perturbations (Seq-DeepFake-P).}

\textcolor{black}{We depict some samples of Seq-DeepFake-P dataset in Fig.~\ref{fig:dataset_perturb}. Total 6 types of manipulations, \textit{e.g.,} JEPG Compression, White Gaussian Noise, Gaussian Blur, Color Saturation Change, Color Contrast Change, and Local Block-Wise Distortion, are applied on all the original Seq-DeepFake images. In particular, JEPG Compression mimics the information loss of images in capture and storage in the real life. White Gaussian Noise, Gaussian Blur and Local Block-Wise Distortion simulate some common visual noise appearing in low-quality images. In addition, Color Saturation Change and Color Contrast Change represent some typical color distortions in images. As illustrated in Fig.~\ref{fig:dataset_perturb}, 1$\sim$4 types of the above perturbations are randomly selected to be added or mixed on samples. Each perturbation is randomly applied by 1$\sim$3 intensity levels. Meanwhile, to avoid severe distribution imbalance, we guarantee each perturbation type does not co-occur for each sample. All of these combinations contribute to in total 360 distinct perturbations with different intensity levels in Seq-DeepFake-P dataset. The variety and comprehensiveness of perturbations facilities the diversity and realness of Seq-DeepFake-P dataset.}
\section{SeqFakeFormer}
\begin{figure}[t] 
	\begin{center}
		\includegraphics[ width=1\linewidth]{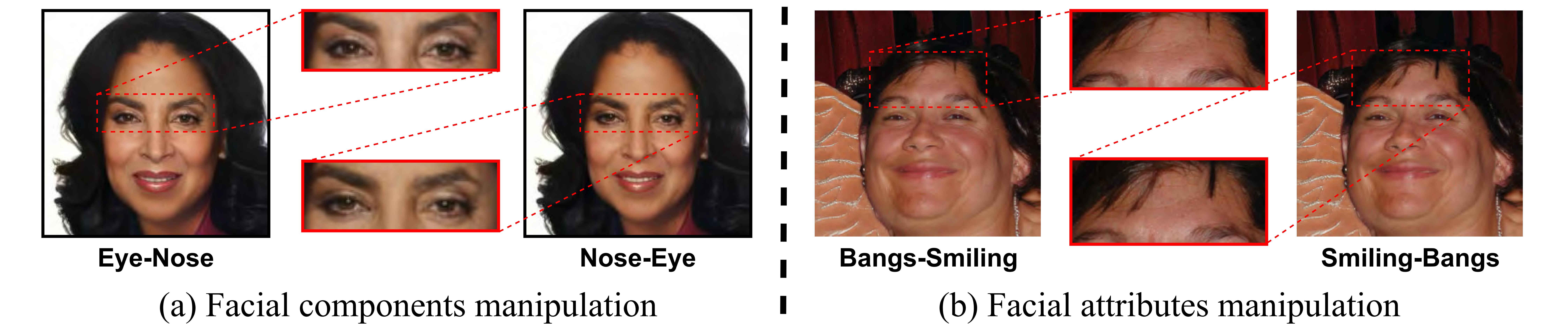}
	\end{center}
	\caption{Effect of different sequential order for facial manipulation. Switching the sequential order of manipulations between (a) eye and nose and (b) bangs and smiling results in different facial manipulations.}
	\label{fig:order}
\end{figure}
\subsection{Motivation}
Most of current facial manipulation applications are constructed using algorithms of Generative Adversarial Network (GAN). However, it is a well known fact that due to imperfect semantic disentanglement in the latent space of GAN~\cite{shen2020interpreting,lee2020high}, manipulating one facial component or attribute is likely to affect the others. As shown in the first row of Fig.~\ref{fig:DatasetIntro}, manipulating the nose in the step of `Eye-Nose' simultaneously results in some little modification on the eye and mouth components compared to the former step `Eye', which alters the overall \textbf{spatial relation} among facial components. We can thus discover some \textbf{spatial manipulation traces} from the spatial relation. Furthermore, as illustrated in Fig.~\ref{fig:order}, switching the sequential order of manipulations (\textit{e.g.,} manipulation order between eye and nose in (a) and bangs and smiling in (b) of Fig.~\ref{fig:order}) causes different facial manipulation results (\textit{e.g.,} distinct gazes in (a) and distinct amount of bangs in (b) in Fig.~\ref{fig:order}). This indicates once changing the sequential order of manipulations, the above overall spatial relation of facial components altered by manipulations will also be changed. This means there exists sequential information from spatial relation that reflects the sequential order of manipulations, which corresponds to the facial manipulation sequence. That is, we can extract the spatial relation among facial components to unveil the \textbf{spatial manipulation traces} and model their \textbf{sequential relation} to detect the facial manipulation sequence. We thus regard the sequential relation as \textbf{sequential manipulation traces}.
\begin{figure*}[t] 
	\begin{center}
		\includegraphics[width=0.9\linewidth]{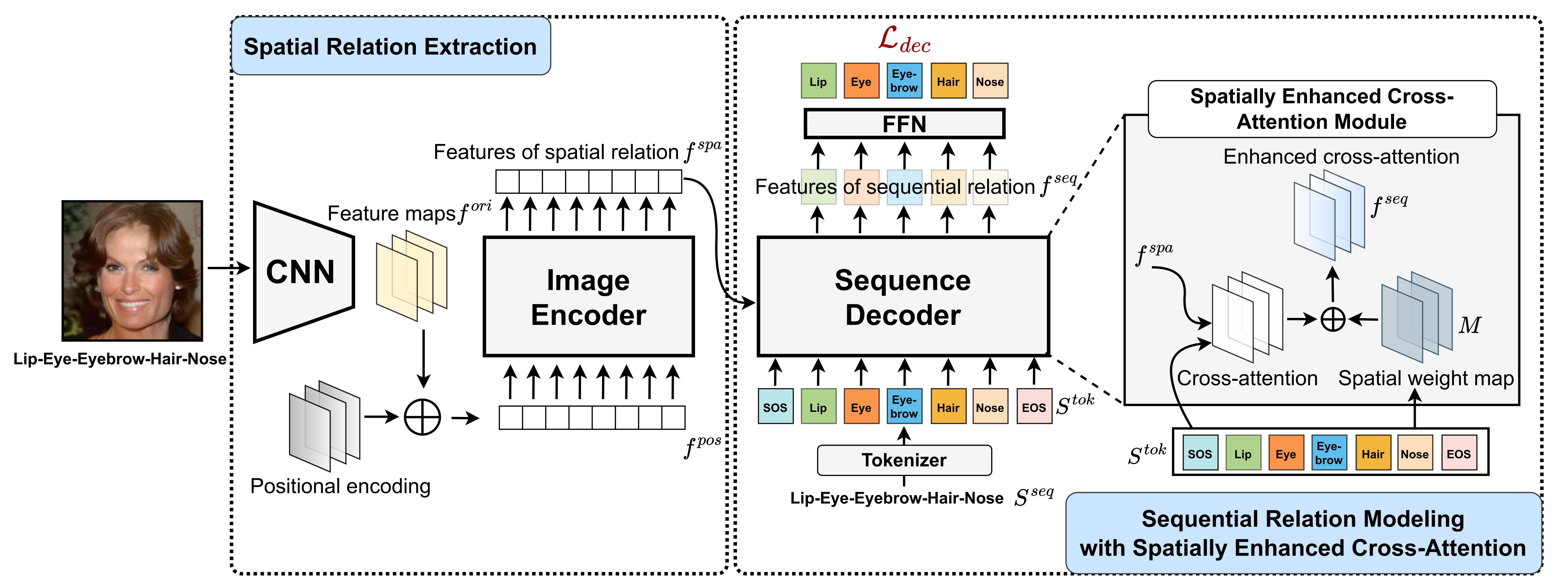}
	\end{center}
	\caption{Overview of proposed Seq-DeepFake Transformer (\textbf{SeqFakeFormer}). We first feed the face image into a CNN to learn features of spatial manipulation regions, and extract their spatial relation via self-attention modules in Image Encoder. Then we model the sequential relation of spatial relation features to detect the manipulation sequences via Sequence Decoder. A spatial enhanced cross-attention module is integrated into the decoder, contributing to a more effective cross-attention.}
	\label{fig:transformer}
	\vspace{-12pt}
\end{figure*}
\subsection{Overview}
Based on the above observation, we cast detecting Seq-DeepFake manipulation as a specific image-to-sequence task, where inputs are manipulated/original images and outputs are facial manipulation sequences. Three challenges will be encountered when addressing this task: \textbf{1)} From Fig.~\ref{fig:DatasetIntro} to \ref{fig:DatasetExp-appendix}, it can be seen that distinguishing original faces and sequential manipulated faces is extremely hard. Besides, with respect to different people, differences in face contour cause diverse manipulation regions for the same type of facial components/attributes manipulation. Thus, given indistinguishable and diverse facial manipulations, how to adaptively capture subtle manipulation regions and model their spatial relation accurately is quite challenging; \textbf{2)} Based on the spatial relation of manipulated components/attributes, how to precisely model their sequential relation so as to detect the sequential facial manipulation is another challenge; \textbf{3)} Compared to normal image-to-sequence task (\textit{e.g.} image captions), the annotations of manipulation sequences are much shorter and thus less informative in our task. Therefore, how to effectively learn the sequential information of facial manipulations given limited manipulation sequences should also be considered.

To cope with the above three challenges, as shown in Fig.~\ref{fig:transformer}, we propose a  Seq-DeepFake  Transformer (\textbf{SeqFakeFormer}), which is composed of two key parts: \textbf{Spatial Relation Extraction} via Image Encoder, \textbf{Sequential Relation Modeling with Spatially Enhanced Cross-Attention} via Sequence Decoder. To capture spatial manipulation traces, features of subtle manipulation regions are first adaptively captured by a CNN and their spatial relation are extracted via self-attention modules in Image Encoder. After that, to capture sequential manipulation traces, we model the sequential relation of spatial relation features through cross-attention layers in Sequence Decoder with an auto-regressive mechanism. To achieve more effective cross-attention given limited annotations of manipulation sequences, a spatially enhanced cross-attention module is devised to generate the adaptive spatial weight map for each corresponding manipulation in the sequence. In following subsections, we describe all components in detail.
\subsection{Spatial Relation Extraction}
\label{sec:SpatialRelation}
To adaptively capture subtle and various facial manipulation regions, we exploit a CNN to learn feature maps of the input image. Given an input image $x$ $\in$ R$^{3 \times H' \times W'}$ from Seq-DeepFake dataset $U$, we first feed it into a CNN~\cite{he2016deep} to extract its visual feature maps $f^{ori}$ = CNN($x$), $f^{ori} \in$ R$^{C \times  H \times W}$, where $H', W'$, and $H, W$ are the height and width of the input image and its corresponding feature maps, respectively. $C$ is the number of channels of feature maps.

Since the transformer architecture is permutation-invariant, we supplement original visual features maps $f^{ori}$ with fixed positional encodings~\cite{parmar2018image,bello2019attention}, resulting in feature maps denoted as $f^{pos}$. Given that a sequence is accepted as input for transformer-based Image Encoder, we reshape the spatial dimensions of $f^{pos}$ to one dimension, generating reshaped features $f^{pos} \in$ R$^{C \times HW}$. After fed into Image Encoder, $f^{pos}$ conducts self-attention by generating the key, query, and value features $K, Q, V$ so as to extract the relation among all spatial positions. Through this self-attention operation on CNN features, spatial relation of manipulation regions are exploited and thus spatial manipulation traces can be extracted. To further facilitate spatial relation extraction, this paper adopts multi-head self-attention which splits features $f^{pos}$ into multiple groups along the channel dimension. The multi-head normalized attention based on dot-product is as follows:
\begin{equation}
\begin{split}
& f^{spa}_i = \text{Softmax}(K_i^T Q_i / \sqrt{d})V_i, f^{spa} = \text{Concat}(f^{spa}_1, ... , f^{spa}_D)
\end{split}
\end{equation}

\noindent where $K_i , Q_i , V_i$ denote the $i$-th group of the key, query, and value features, $d$ is dimension of queries and keys, and total $D$ groups are generated. We then concatenate all the groups to form the features of spatial relation $f^{spa}$ as the output of encoder.

\begin{figure*}[t] 
	\begin{center}
		\includegraphics[width=1\linewidth]{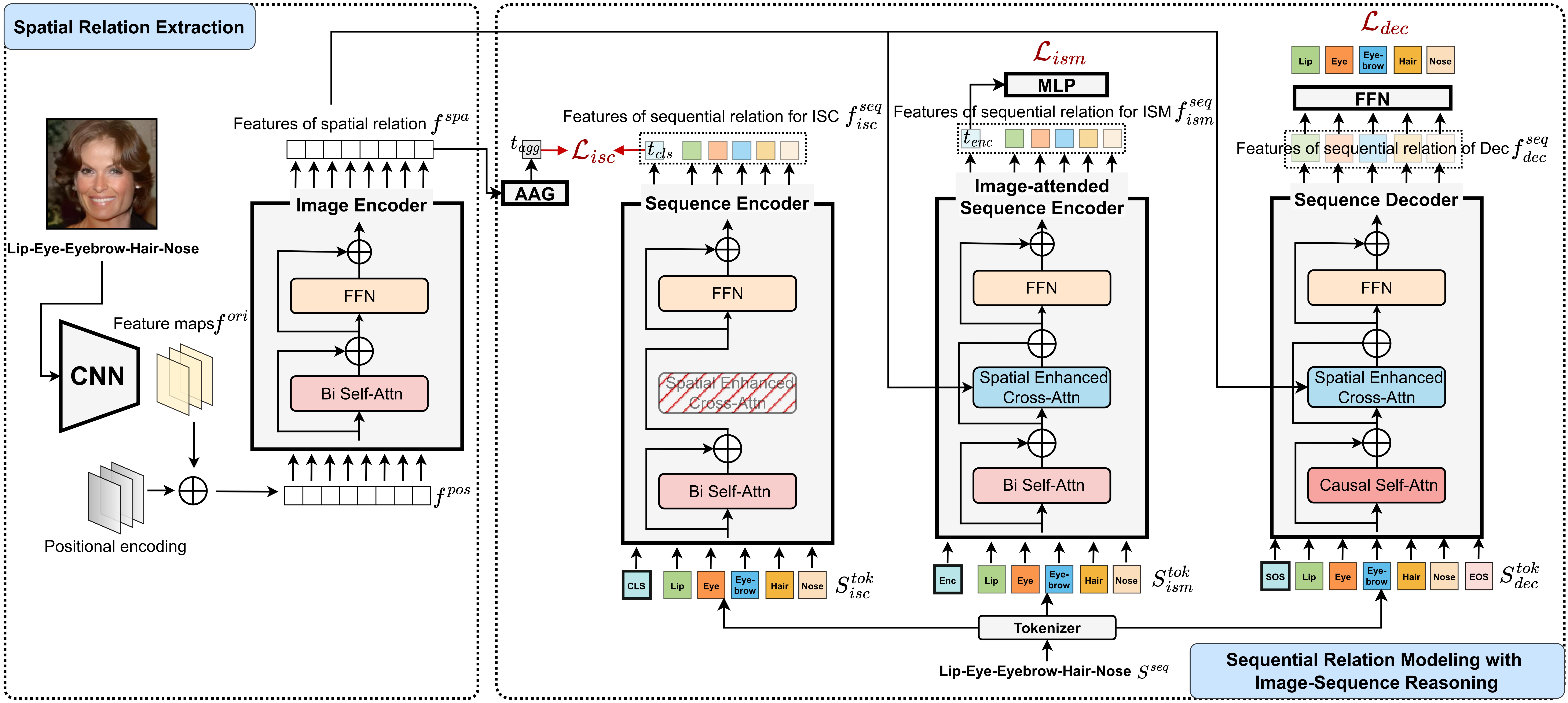}
	\end{center}
	\caption{\textcolor{black}{Overview of proposed Seq-DeepFake Transformer with Image-Sequence Reasoning (\textbf{SeqFakeFormer++}). We further integrate two more Image-Sequence Reasoning modules, \textit{i.e.,} Sequence Encoder with Image-Sequence Contrastive Learning, and Image-attended Sequence Encoder with Image-Sequence Matching into our model. This contributes to deeper reasoning of image-sequence relation and thus more robust detection of Seq-DeepFake under various perturbations.}}
	\label{fig:seqfakeformer++}
\end{figure*}

\subsection{Sequential Relation Modeling with Spatially Enhanced Cross-Attention}

Given features of spatial relation $f^{spa}$ extracted from the encoder, we propose to model the sequential relation among them to detect the facial manipulation sequences. To this end, we carry out cross-attention between features of spatial relation $f^{spa}$ and corresponding annotations of manipulation sequences in an auto-regressive manner. To achieve this, we send manipulation sequences $S^{ori} \in R^{C \times N}$ (\textit{e.g.,} $N$=5 in Fig.~\ref{fig:transformer} before a Tokenizer) into a Tokenizer, where we transform each manipulation in the sequence into one token and insert Start of Sentence \texttt{[SOS]} and End of Sentence \texttt{[EOS]} tokens into the beginning and end of sequence. After that, we obtain tokenized manipulation sequences $S^{tok} \in R^{C \times (N+2)}$ to be cross-attended with features of spatial relation $f^{spa}$. Base on the auto-regressive mechanism, the decoding process of manipulation sequence in Sequence Decoder (aided by cross-attention) is triggered by \texttt{[SOS]} token and will be automatically stopped once \texttt{[EOS]} token is predicted. In this way, we can predict facial manipulation sequences with adaptive lengths.

Normally, cross-attention between tokenized sequences $S^{tok}$ and features of spatial relation $f^{spa}$ should be performed directly. However, as mentioned above, compared to the normal image-to-sequence task, annotations of manipulation sequences are much shorter and thus less informative ($S^{tok}$ only has $(N+2)$-length and maximum of $N$ is 5). To effectively cross-attend features of spatial relation with limited manipulation sequences, inspired by~\cite{gao2021fast}, we propose a sequential relation modeling with Spatially Enhanced Cross-Attention (SECA). We argue that each manipulation in $S^{tok}$ corresponds to one specific facial component/attribute which has a strong prior of spatial regions, thus we can enrich the information of manipulation sequences guided by this prior. To this end, we generate the spatial weight map for each manipulation by dynamically predicting the spatial center and scale of each manipulation component/attribute in sequences by:

\begin{equation}
\begin{split}
& t_h, t_w =  \text{sigmoid}(\text{MLP}(S^{tok})), r_h, r_w = \text{FC}(S^{tok}) \\
\end{split}
\label{equ:centerscale}
\end{equation}

\noindent where $t_h, t_w$ and $r_h, r_w$ are estimated 2-dimensional coordinates corresponding to spatial centers and scales of specific manipulations in the sequences, respectively. Then the Gaussian-shape spatial weight map can be generated as:

\begin{equation}
\begin{split}
& M(h, w) =  \text{exp}\left(-\frac{(h-t_h)^2}{\lambda r_h^2}-\frac{(w-t_w)^2}{\lambda r_w^2} \right)
\end{split}
\label{equ:spatialmap}
\end{equation}

\noindent where $(h, w) \in [0, H] \times [0, W]$ are 2-dimensional coordinates of the spatial weight map $M$, and $\lambda$ is a hyper-parameter to modulate the bandwidth of the Gaussian-shape distribution. From Eq.~\ref{equ:spatialmap}, it can be seen that spatial weight map $M$ can assign higher importance to spatial regions closer to the centers and lower weights to locations farther from the centers. Moreover, as analyzed before, since diverse manipulation regions are presented for different people, the above dynamically learned scales can further tune the height/width ratios of spatial weight map based on each manipulation, contributing to a more adaptive spatial weight map. Based on this idea, we can enhance cross-attention between features of spatial relation and manipulation sequences with generated spatial weight map $M$ as follows:

\begin{equation}
\begin{split}
& S = \text{FC}(S^{tok}), K, V = \text{FC}(f^{spa}), \\
& f^{seq}_i = \text{Softmax}(K_i^T Q_i/ \sqrt{d} + logM)V_i, \\
& f^{seq} = \text{Concat}(f^{seq}_1, ..., f^{seq}_D)
\end{split}
\label{equ:decoder}
\end{equation}

\noindent where FC denotes a single fully-connected layer, and $f^{seq}_i$ denotes features of sequential relation. The cross-attention of the $i$-th head is further element-wise added with logarithm of spatial weight map $M$, which contributes to spatially enhanced cross-attention. Furthermore, to model the sequential relation of facial manipulation, the auto-regressive mechanism is integrated into the above cross-attention process. This is implemented by masking out (setting to $-\infty$) all values in the input of the Softmax function in Eq.~\ref{equ:decoder} which correspond to illegal connections. Through concatenation of features of sequential relation from all cross-attention heads, we can obtain the final features of sequential relation $f^{seq}$ as the output of decoder.

\textcolor{black}{The features of sequential relation are then fed into a Fast Forward Network (FFN) to be predicted as a class score for each manipulation. Finally, we jointly train the CNN, Image Encoder and Sequence Decoder by minimizing cross-entropy loss between each class score and corresponding annotation in the manipulation sequence as follows,}
\textcolor{black}{\begin{equation}
\vspace*{-1mm}
\begin{split}
\mathcal{L}_{dec} = \mathbb{E}_{(x,S^{seq})\sim U}{\rm \textbf{H}}({\rm FFN}(f^{seq}), S^{seq})
\end{split}
\end{equation}}
\textcolor{black}{where ${\rm \textbf{H}(\cdot)}$ is the cross-entropy function. }

\textcolor{black}{\subsection{Multi-head version of SECA}}
\textcolor{black}{Similar to~\cite{gao2021fast}, we extend the basic version of Spatially Enhanced Cross-Attention (SECA) into multi-heads version, enhancing cross-attention features adaptively for different cross-attention heads. As formulated in Eq.~\ref{equ:centerscale}, the basic version of SECA estimates 2-dimensional coordinates corresponding to spatial centers [$t_h, t_w$]. Similarly, the multi-head version of SECA estimates a head-shard spatial center [$t_h, t_w$] and then predicts a head-specific center offset [$\triangle t_{h,i}, \triangle t_{w,i}$] and corresponding head-specific scales [$r_{h,i}, r_{w,i}$] for $i$-th cross-attention head. In this way, we generate $i$-th head-specific Gaussian-shape spatial weight map $M_i$ based on the $i$-th head-specific center [$t_h + \triangle t_{h,i}, t_w + \triangle t_{w,i}$] and scales [$r_{h,i}, r_{w,i}$] as:}
\textcolor{black}{\begin{equation}
\footnotesize
\begin{split}
& M_i(h, w) =  \text{exp}\left(-\frac{(h-(t_h + \triangle t_{h,i}))^2}{\lambda r_{h,i}^2}-\frac{(w-(t_w + \triangle t_{w,i}))^2}{\lambda r_{w,i}^2} \right)
\end{split}
\label{equ:spatialmap-appendix}
\end{equation}}
\textcolor{black}{\noindent Based on this, we can calculate the features of sequential relation $f^{seq}_i$ from $i$-th cross-attention head enhanced by the $i$-th SECA as follows:}
\textcolor{black}{\begin{equation}
\begin{split}
& f^{seq}_i = \text{Softmax}(K_i^T Q_i /\sqrt{d} + logM_i)V_i, \\
\end{split}
\label{equ:decoder-appendix}
\end{equation}}
\textcolor{black}{\noindent Different from basic version of SECA, Eq.~\ref{equ:decoder-appendix} shows that in the multi-head version of SECA, the cross-attention of the $i$-th head is element-wise added with logarithm of $i$-th head-specific spatial weight map $M_i$, which contributes to a more adaptive and specific enhanced cross-attention.}

\textcolor{black}{\section{SeqFakeFormer++}}
\textcolor{black}{Although SeqFakeFormer can perform well in Seq-DeepFake dataset, its performance is prone to degenerate when facing Seq-DeepFake data with perturbations in Seq-DeepFake-P. This is because SeqFakeFormer is only able to exploit the shallow relation between images and sequences merely via cross-attention between Image Encoder and Sequence Decoder. This shallow relation becomes less robust when dealing with data in Seq-DeepFake-P dataset as perturbations could cover or alter some manipulation traces. To more robustly detect sequential manipulation sequences under various post-processing perturbations, we further devise a Seq-DeepFake Transformer with Image-Sequence Reasoning (\textbf{SeqFakeFormer++}) as illustrated in Fig.~\ref{fig:seqfakeformer++}. It complements two more Image-Sequence Reasoning modules, \textit{i.e.,} Sequence Encoder and Image-attended Sequence Encoder, to explore deeper image-sequence relation. This strengthens the correspondence between image-sequence pairs and thus improves the robustness of Seq-DeepFake detection to various noises.}

\textcolor{black}{\subsection{Image-Sequence Contrastive Learning}} 
\textcolor{black}{In essence, feature distances between images and corresponding manipulation sequences should closer to those of unmatched pairs. To make our model more robust to perturbations, this image-sequence correlation should be further strengthened. To this end, we propose Image-Sequence Contrastive Learning to align the uni-modal feature representations of images and sequences. As depicted in Fig.~\ref{fig:seqfakeformer++}, we exploit the features of spatial relation $f^{spa}$ for the uni-modal feature representations of images. Specifically, to adaptively aggregate all the feature tokens in $f^{spa}$, we propose Attentional Aggregation (\textbf{AAG}) to learn a global feature representation for the whole image via an attention mechanism. To be specific, \textbf{AAG} is performed by cross-attending a $t_{agg}$ token with all the tokens in $f^{spa}$:}

\textcolor{black}{\begin{equation}
\begin{split}
&t_{agg} = \textbf{AAG}(f^{spa})= \text{Attention}(t_{agg}, f^{spa}, f^{spa}) \\
\end{split}
\end{equation}}
\textcolor{black}{where $\text{Attention}(Q, K, V) =\text{Softmax}(K^T Q / \sqrt{D})V$, and $t_{agg}$ serves as the feature representation of images.}

\textcolor{black}{As for learning uni-modal feature representations of sequences, we add one more Sequence Encoder as shown in Fig.~\ref{fig:seqfakeformer++}, where Bi-directional Self-Attention layers are built without Spatial Enhanced Cross-Attention layers (SECA). We send original annotations of manipulation sequences $S^{ori}$ into the Tokenizer which appends a \texttt{[CLS]} token into the beginning of the sequence input to summarize its information. This generates a tokenized manipulation sequence for Image-Sequence Contrastive Learning denoted as $S^{tok}_{isc}$. We feed it into Sequence Encoder and obtain the corresponding features of sequential relation for ISC $f^{seq}_{isc}$, where the first feature token is $t_{cls}$ serving as the feature representation of whole sequence. Based on the feature representation of images $t_{agg}$ and sequences $t_{cls}$, we perform ISC by pulling features of matched image-sequence pairs close while pushing those of unmatched pairs apart. Following InfoNCE loss~\cite{oord2018representation}, we formulate image-to-sequence contrastive loss by:}
\textcolor{black}{\begin{equation}
\begin{split}
\mathcal{L}_{i2s}(I, S^{+}, S^{-}) = -\mathbb{E}_{p(I,S)}\left[ {\rm log} \frac{{\rm exp}(F(I, S^{+})/\tau)}{\sum\nolimits_{k=1}^{K} {\rm exp} (F(I, S^{-}_{k})/\tau)} \right]
\end{split}
\end{equation}}
\textcolor{black}{where $\tau$ is a temperature hyper-parameter, $S^{-} = \{ S^{-}_1, ..., S^{-}_K\}$ is a set of negative sequence samples that are not matched to $I$ in the current batch ($K$ is batch size). We use two projection heads $h_i$ and $h_s$ to map feature representations of two modalities to a lower-dimensional (256) embedding space for similarity calculation: $F(I, S)=h_i(t_{agg})^{\rm T}h_s(t_{cls})$. Similarly, sequence-to-image contrastive loss is as follows:}
\textcolor{black}{\begin{equation}
\begin{split}
\mathcal{L}_{s2i}(S, I^{+}, I^{-}) = -\mathbb{E}_{p(I,S)}\left[ {\rm log} \frac{{\rm exp}(F(S, I^{+})/\tau)}{\sum\nolimits_{k=1}^{K} {\rm exp} (F(S, I^{-}_{k})/\tau)} \right]
\end{split}
\end{equation}}
\textcolor{black}{where $I^{-} = \{ I^{-}_1, ..., I^{-}_K\}$ is a queue of $K$ negative image samples that are not matched to $S$. We incorporate the above two losses and form Image-Sequence Contrastive Loss as:}
\textcolor{black}{\begin{equation}
\begin{split}
\mathcal{L}_{isc} = \frac{1}{2} [\mathcal{L}_{i2s}(I, S^{+}, S^{-}) + \mathcal{L}_{s2i}(S, I^{+}, I^{-}) ]
\end{split}
\end{equation}}
 
\textcolor{black}{\subsection{Image-Sequence Matching}}
\textcolor{black}{To capture more fine-grained alignment between images and sequences, we further interact image features with sequence features to learn image-sequence multi-modal representation. In this regard, we integrate one more Image-attended Sequence Encoder as displayed in Fig.~\ref{fig:seqfakeformer++}, where interaction of cross-modal information is performed by inserting SECA layers the same as those of Sequence Decoder. To explore the correspondence regarding global feature representation between images and sequences, unlike auto-regressive mechanism performed in Causal Self-attention layers of Sequence Decoder, Bi-directional Self-Attention layers are adopted in Image-attended Sequence Encoder. Notably, a task-specific \texttt{[Enc]} token is appended to the beginning of original sequence, which plays the role of the information summary of whole input sequence $S_{ism}^{tok}$. After cross-modal interaction through Image-attended Sequence Encoder, a token $t_{enc}$ can be generated in the output feature $f_{ism}^{feq}$ which corresponds to the input \texttt{[Enc]} token. In this way, $t_{enc}$ summarizes the information of cross-modal interaction and thus can be used as the multi-modal representation of each image-sequence pair. Based on it, we perform ISM by predicting binary classes to determine whether the input image-sequence pair is matched or unmatched. To achieve this, we feed $t_{enc}$ into a Multilayer Perceptron (MLP) and calculate Image-Sequence Matching Loss as follows,}
\textcolor{black}{\begin{equation}
\vspace*{-3mm}
\begin{split}
\mathcal{L}_{ism} = \mathbb{E}_{(I,S)\sim P}{\rm \textbf{H}}({\rm MLP}(t_{enc}), y_{mat})
\end{split}
\end{equation}}
\textcolor{black}{where $y_{mat} \in \{0, 1 \}$ describes whether the input image-sequence pair is matched or not. In order to find more informative unmatched samples, inspired by \cite{li2021align}, we deploy the hard negative mining strategy based on contrastive similarity calculated in Image-Sequence Contrastive Learning. The hardest unmatched samples are selected based on the highest contrastive similarity in a batch.}

\textcolor{black}{Based on the aforementioned analysis, we formulate the above objectives into a unified joint optimization model for SeqFakeFormer++ as follows:}
\textcolor{black}{\begin{equation}
\begin{split}
\mathcal{L} = \mathcal{L}_{isc} + \mathcal{L}_{ism} + \mathcal{L}_{dec}
\end{split}
\label{equ:opt++} 
\end{equation}}

\textcolor{black}{Since SECA and FFN have similar functions in sequence encoder and decoder, we thus share the parameters of these layers when we perform the above joint optimization. In this way, the benefit of deeper Image-Sequence Reasoning brought by ISC and ISM directly contributes to better SECA and FFN layers in Sequence Decoder. This consequently facilitates the auto-regressive sequence decoding carried out in Causal Self-attention layers in Sequence Decoder. All these modules together contribute to a more robust and accurate detection of Seq-DeepFake manipulation sequences even facing perturbations.}

\section{Experiments}
\subsection{Experimental Setup}
\noindent \textbf{Implementation Details.} 
We choose two different CNNs, ResNet-34~\cite{he2016deep} and ResNet-50~\cite{he2016deep} pre-trained on ImageNet~\cite{deng2009imagenet} dataset in our paper. To be comparable in the number of parameters, we adopt a transformer model with 2  encoder and 2 decoder layers with 4 attention heads. For the training schedule, we employ 20 epochs warm-up strategy and train for 170 epochs with a learning rate drop to 10\% in every 50 epochs. The initial learning rates are set as 1$e-3$ for transformer part and 1$e-4$ for CNN part. We set $\lambda=4$.

\textcolor{black}{Implementation is in PyTorch. For the training schedule, we employ a 20-epochs warm-up strategy. The initial learning rate is set as 1$e-3$ for transformer part and 1$e-4$ for CNN part, with a decay factor of 10 at 70 and 120 epochs, totally 170 epochs. We use the SGD momentum optimizer with weight decay set as $1e-4$. We use a mini-batch size of 32 per GPU and 4 GPUs in total. Model selection for evaluation is conducted by considering the trained model that has produced the best accuracy on the validation set.}

\noindent \textbf{Baseline Methods.} 
The most straightforward solution for detecting Seq-DeepFake manipulation is to regard it as a multi-label classification problem~\cite{wang2021can}. It treats all manipulations in the sequences as independent classes and classifies the manipulated images into multiple manipulation classes. Specifically, we design a simple multi-label classification network (denoted as \textbf{Multi-Cls}) as one of the baselines. We use ResNet-34~\cite{he2016deep} and ResNet-50~\cite{he2016deep} pre-trained on ImageNet~\cite{deng2009imagenet} dataset as backbones for the multi-label classification network, which is concatenated with $N$ single linear-layer branches as $N$ classification heads ($N=5$ as maximum manipulation steps are 5 in Seq-DeepFake dataset). Moreover, we study a more complex transformer structure modified for our problem. \textbf{DETR}~\cite{carion2020end} is a popular transformer devised for end-to-end object detection. This model detects input images' bounding boxes and corresponding object classes conditioned on object queries. We revise this model by replacing the object queries with annotations of manipulation sequences and only preserve the output of object classes. 

Furthermore, to examine the performance of existing deepfake detection methods for our research problem, we adapt three state-of-the-art deepfake detection methods, a Dilated Residual Network variant (\textbf{DRN})~\cite{wang2019detecting}, a two-stream network modeling the correlation between high-frequency features and regular RGB features (\textbf{TS})~\cite{luo2021generalizing}, and a multi-attentional deepfake detection (\textbf{MA})~\cite{zhao2021multi}, into multi-label classification setting. To be specific, we replace their binary label classifier with multiple classification heads to classify sequential manipulations. Please note since all of the above baselines are only able to predict the facial manipulation with fixed length ($N=5$), `no manipulation' class will be padded into the annotation sequence if its length is shorter than $N$ so that we can keep the same length between predictions and annotation sequences for training. 

\noindent \textbf{Evaluation Metrics.} We propose two evaluation metrics for this new task.

\begin{figure}[t] 
	\begin{center}
		\includegraphics[width=1\linewidth]{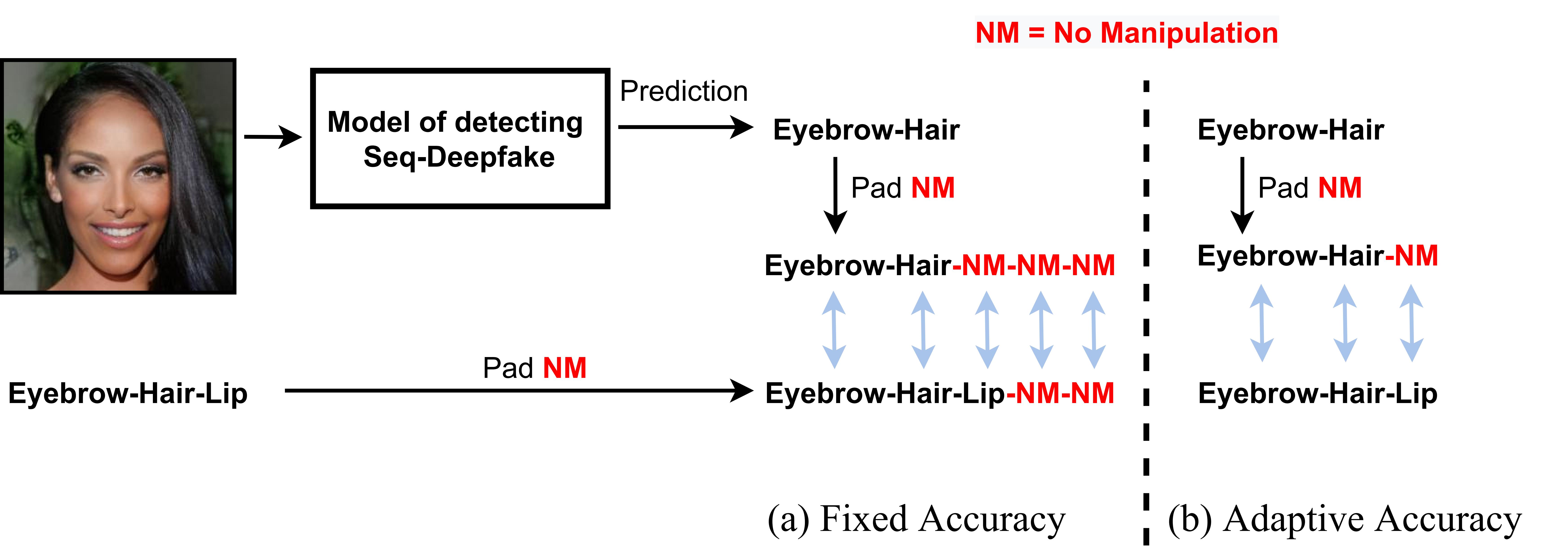}
	\end{center}
	\caption{\textcolor{black}{Comparison between two evaluation metrics (a) Fixed Accuracy and (b) Adaptive Accuracy.}}
	\label{fig:metric}
\end{figure}

\begin{table}[t]
\scriptsize
\renewcommand{\arraystretch}{1}
\centering
\caption{Accuracy of detecting Seq-DeepFake based on sequential facial components manipulation.}
\begin{tabular}{c|cc|cc}
\Xhline{2.5\arrayrulewidth}
\multirow{2}{*}{Methods} & \multicolumn{2}{c|}{ResNet-34}                       & \multicolumn{2}{c}{ResNet-50}                        \\ \cline{2-5} 
                         & \multicolumn{1}{c|}{Fixed-Acc}      & Adap-Acc   & \multicolumn{1}{c|}{Fixed-Acc}      & Adap-Acc   \\ \hline
Multi-Cls                & \multicolumn{1}{c|}{69.66}          & 50.54          & \multicolumn{1}{c|}{69.65}          & 50.57          \\
DETR~\cite{carion2020end}                     & \multicolumn{1}{c|}{69.87}          & 50.63          & \multicolumn{1}{c|}{69.75}          & 49.84          \\
Ours                     & \multicolumn{1}{c|}{72.13} & 54.80 & \multicolumn{1}{c|}{72.65} & 55.30 \\ 

Ours++ & \multicolumn{1}{c|}{\textcolor{black}{\textbf{72.66}}} & \textcolor{black}{\textbf{55.16}} & \multicolumn{1}{c|}{\textcolor{black}{\textbf{72.81}}} & \textcolor{black}{\textbf{56.12}} \\
\Xhline{2.5\arrayrulewidth}
\end{tabular}
\label{tbl:facecomp}
\end{table}

\begin{table}[t]
\scriptsize
\renewcommand{\arraystretch}{1}
\centering
\caption{Accuracy of detecting Seq-DeepFake based on sequential facial attributes manipulation.}
\begin{tabular}{c|cc|cc}
\Xhline{2.5\arrayrulewidth}
\multirow{2}{*}{Methods} & \multicolumn{2}{c|}{ResNet-34}                       & \multicolumn{2}{c}{ResNet-50}                        \\ \cline{2-5} 
                         & \multicolumn{1}{c|}{Fixed-Acc}      & Adap-Acc   & \multicolumn{1}{c|}{Fixed-Acc}      & Adap-Acc   \\ \hline
Multi-Cls                & \multicolumn{1}{c|}{66.99}          & 46.68          & \multicolumn{1}{c|}{66.66}          & 46.00          \\
DETR~\cite{carion2020end}                     & \multicolumn{1}{c|}{67.93}          & 48.15          & \multicolumn{1}{c|}{67.62}          & 47.99          \\
Ours                     & \multicolumn{1}{c|}{67.99} & 48.32 & \multicolumn{1}{c|}{68.86} & 49.63 \\ 
Ours++ & \multicolumn{1}{c|}{\textcolor{black}{\textbf{68.74}}} & \textcolor{black}{\textbf{49.51}} & \multicolumn{1}{c|}{\textcolor{black}{\textbf{69.03}}} & \textcolor{black}{\textbf{49.92}} \\
\Xhline{2.5\arrayrulewidth}
\end{tabular}
\label{tbl:faceattr}
\end{table}

\begin{table}[t]
\scriptsize
\renewcommand{\arraystretch}{1}
\centering
\caption{Accuracy of detecting Seq-DeepFake compared to deepfake detection methods.}
\begin{tabular}{c|cc|cc}
\Xhline{2.5\arrayrulewidth}
\multirow{2}{*}{Methods} & \multicolumn{2}{c|}{Face Components Manipulation}    & \multicolumn{2}{c}{Face Attributes Manipulation}     \\ \cline{2-5} 
                         & \multicolumn{1}{c|}{Fixed-Acc}      & Adap-Acc   & \multicolumn{1}{c|}{Fixed-Acc}      & Adap-Acc   \\ \hline
DRN~\cite{wang2019detecting}               & \multicolumn{1}{c|}{66.06}          & 45.79          & \multicolumn{1}{c|}{64.42}          & 43.20          \\
MA~\cite{zhao2021multi}                     & \multicolumn{1}{c|}{71.31}          & 52.94          & \multicolumn{1}{c|}{67.58}          & 47.48      \\
TS~\cite{luo2021generalizing}               & \multicolumn{1}{c|}{71.92}          & 53.89          & \multicolumn{1}{c|}{66.77}          & 46.38          \\
Ours                     & \multicolumn{1}{c|}{72.65} & 55.30 & \multicolumn{1}{c|}{68.86} & 49.63 \\ 
Ours++ & \multicolumn{1}{c|}{\textcolor{black}{\textbf{72.81}}} & \textcolor{black}{\textbf{56.12}} & \multicolumn{1}{c|}{\textcolor{black}{\textbf{69.03}}} & \textcolor{black}{\textbf{49.92}} \\
\Xhline{2.5\arrayrulewidth}
\end{tabular}
\label{tbl:deepfakebaseline}
\end{table}

\begin{itemize}[leftmargin=*]
\item \textbf{Fixed Accuracy (Fixed-Acc):} \textcolor{black}{Since all the baseline methods are only able to predict the facial manipulation with fixed length ($N=5$), `no manipulation' class will be padded into the annotation sequence if its length is shorter than $N$ so that we can keep the same length between predictions and annotation sequences for training. Following this strategy, as shown in Fig.~\ref{fig:metric}, under the evaluation metric of Fixed Accuracy, given the model prediction, such as `Eyebrow-Hair', we first pad `no manipulation' class into it to form the padded prediction sequence as `Eyebrow-Hair-NM-NM-NM' (NM means `no manipulation' class), generating the prediction with fixed $N$-length ($N=5$). To keep the same length between predictions and annotation sequences for evaluation, we pad `no manipulation' class into the annotation of manipulation sequences as well, denoted as `Eyebrow-Hair-Lip-NM-NM'. Given this, we compare each manipulation class in the predicted sequences with its corresponding annotation to calculate the evaluation accuracy.}

\item \textbf{Adaptive Accuracy (Adaptive-Acc/Adap-Acc):} \textcolor{black}{Moreover, since the proposed method detects facial manipulation sequences in auto-regressive mechanism, detection will be automatically stopped once predicting the \texttt{[EOS]} token. As such, the proposed method can detect facial manipulation sequences with adaptive lengths. To conduct the evaluation in this scenario, as illustrated in Fig.~\ref{fig:metric}, the second type of evaluation is devised, which compares predicted manipulations and corresponding annotations within the maximum steps of manipulations ($N = 3$ in Fig.~\ref{fig:metric} and we just pad one `no manipulation' class into prediction sequence) between them. This makes the evaluation more concentrated on the accuracy of manipulations.}
\end{itemize}

\subsection{Benchmark for Seq-DeepFake}

We tabulate the first benchmark for detecting sequential facial manipulation based on facial components manipulation and facial attributes manipulation in Tables~\ref{tbl:facecomp} to \ref{tbl:deepfakebaseline}. We note that, both baselines and the proposed method yield much higher performance under evaluation metric Fixed-Acc than Adaptive-Acc. This validates that detecting sequential facial manipulation with adaptive lengths is much harder than its simplified version with fixed length. It can be observed from Tables~\ref{tbl:facecomp} and \ref{tbl:faceattr}, that the proposed SeqFakeFormer obtains the best performance of detecting facial manipulation sequences compared to all considered baselines in both facial components manipulation and facial attributes manipulation. In addition, SeqFakeFormer also performs better than other baselines with both CNNs (ResNet-34 and ResNet-50), indicating the compatibility of the proposed method with different feature extractors. Specifically, the proposed method has achieved about 3-4\% improvement in facial components sequential manipulation and 1-2\% improvement in facial attributes sequential manipulation under two evaluation metrics. In particular, there exists a larger performance gap between SeqFakeFormer and other baselines under evaluation metric Adaptive-Acc than Fixed-Acc, suggesting SeqFakeFormer is more effective in the more challenging case. 

In addition, we tabulate the comparison between three SOTA deepfake detection methods and our method in Table~\ref{tbl:deepfakebaseline}. SeqFakeFormer also outperforms all SOTA deepfake detection methods in both manipulation types. Since all the baselines treat detecting Seq-DeepFake as a multi-label classification problem, only spatial information of manipulated images are extracted. In contrast, SeqFakeFormer is capable of exploiting both spatial and sequential manipulation traces and thus more useful sequential information can be modeled, which is the key to enhance the performance of Seq-DeepFake Detection.

\textcolor{black}{Furthermore, it can be seen from Tables~\ref{tbl:facecomp} to  \ref{tbl:deepfakebaseline} that the upgraded SeqFakeFormer++ (Ours++) also exceeds other baselines and further improves the performance of Seq-FakeFormer under all settings. This demonstrates the effectiveness of Seq-FakeFormer++ on original Seq-DeepFake dataset without perturbations.}

\subsection{Ablation Study}

\begin{table}[t]
\scriptsize
\renewcommand{\arraystretch}{1}
\centering
\caption{Ablation study of detecting Seq-DeepFake based on sequential facial components manipulation.}
\begin{tabular}{cc|cc|cc}
\Xhline{2.5\arrayrulewidth}
\multicolumn{2}{c|}{Components}                      & \multicolumn{2}{c|}{ResNet-34}                       & \multicolumn{2}{c}{ResNet-50}                        \\ \hline
\multicolumn{1}{c|}{Auto-reg} & SECA          & \multicolumn{1}{c|}{Fixed-Acc}      & Adap-Acc   & \multicolumn{1}{c|}{Fixed-Acc}      & Adap-Acc   \\ \hline
\multicolumn{1}{c|}{\XSolidBrush}     & \XSolidBrush   & \multicolumn{1}{c|}{70.64}          & 52.19          & \multicolumn{1}{c|}{71.22}          & 53.43          \\
\multicolumn{1}{c|}{\XSolidBrush}     & \CheckmarkBold & \multicolumn{1}{c|}{70.77}          & 51.71          & \multicolumn{1}{c|}{70.99}          & 52.66          \\
\multicolumn{1}{c|}{\CheckmarkBold}   & \XSolidBrush   & \multicolumn{1}{c|}{71.88}          & 53.84          & \multicolumn{1}{c|}{72.18}          & 54.64          \\
\multicolumn{1}{c|}{\CheckmarkBold}   & \CheckmarkBold & \multicolumn{1}{c|}{\textbf{72.13}} & \textbf{54.80} & \multicolumn{1}{c|}{\textbf{72.65}} & \textbf{55.30} \\ \Xhline{2.5\arrayrulewidth}
\end{tabular}
\label{tbl:Ablfacecomp}
\end{table}

\begin{table}[t]
\scriptsize
\renewcommand{\arraystretch}{1}
\centering
\caption{Ablation study of detecting Seq-DeepFake based on sequential facial attributes manipulation.}
\begin{tabular}{cc|cc|cc}
\Xhline{2.5\arrayrulewidth}
\multicolumn{2}{c|}{Components}                      & \multicolumn{2}{c|}{ResNet-34}                       & \multicolumn{2}{c}{ResNet-50}                        \\ \hline
\multicolumn{1}{c|}{Auto-reg} & SECA          & \multicolumn{1}{c|}{Fixed-Acc}      & Adap-Acc   & \multicolumn{1}{c|}{Fixed-Acc}      & Adap-Acc   \\ \hline
\multicolumn{1}{c|}{\XSolidBrush}     & \XSolidBrush   & \multicolumn{1}{c|}{66.98}          & 45.87          & \multicolumn{1}{c|}{68.14}          & 48.49          \\
\multicolumn{1}{c|}{\XSolidBrush}     & \CheckmarkBold & \multicolumn{1}{c|}{67.36}          & 47.22          & \multicolumn{1}{c|}{68.77}          & 49.54          \\
\multicolumn{1}{c|}{\CheckmarkBold}   & \XSolidBrush   & \multicolumn{1}{c|}{66.70}          & 46.56          & \multicolumn{1}{c|}{68.17}          & 48.81          \\
\multicolumn{1}{c|}{\CheckmarkBold}   & \CheckmarkBold & \multicolumn{1}{c|}{\textbf{67.99}} & \textbf{48.32} & \multicolumn{1}{c|}{\textbf{68.86}} & \textbf{49.63} \\ \Xhline{2.5\arrayrulewidth}
\end{tabular}
\label{tbl:Ablfaceattr}
\vspace{-10pt}
\end{table}

In this sub-section we investigate the impact of two key components in SeqFakeFormer, auto-regressive mechanism and Spatially Enhanced Cross-Attention module (SECA), to the overall performance. The considered components and the corresponding results obtained for each case are tabulated Tables~\ref{tbl:Ablfacecomp} and \ref{tbl:Ablfaceattr}. As evident from Tables~\ref{tbl:Ablfacecomp} and \ref{tbl:Ablfaceattr}, removing either auto-regressive mechanism or SECA will degrade the overall performance. This validates that auto-regressive mechanism facilitates the sequential relation modeling and SECA benefits the cross-attention. These components complement each other to produce better performance for detecting Seq-DeepFake.

\textcolor{black}{\subsection{Accuracy for Each Manipulation Sequence}}

\begin{figure}[t] 
	\begin{center}
		\includegraphics[width=1\linewidth]{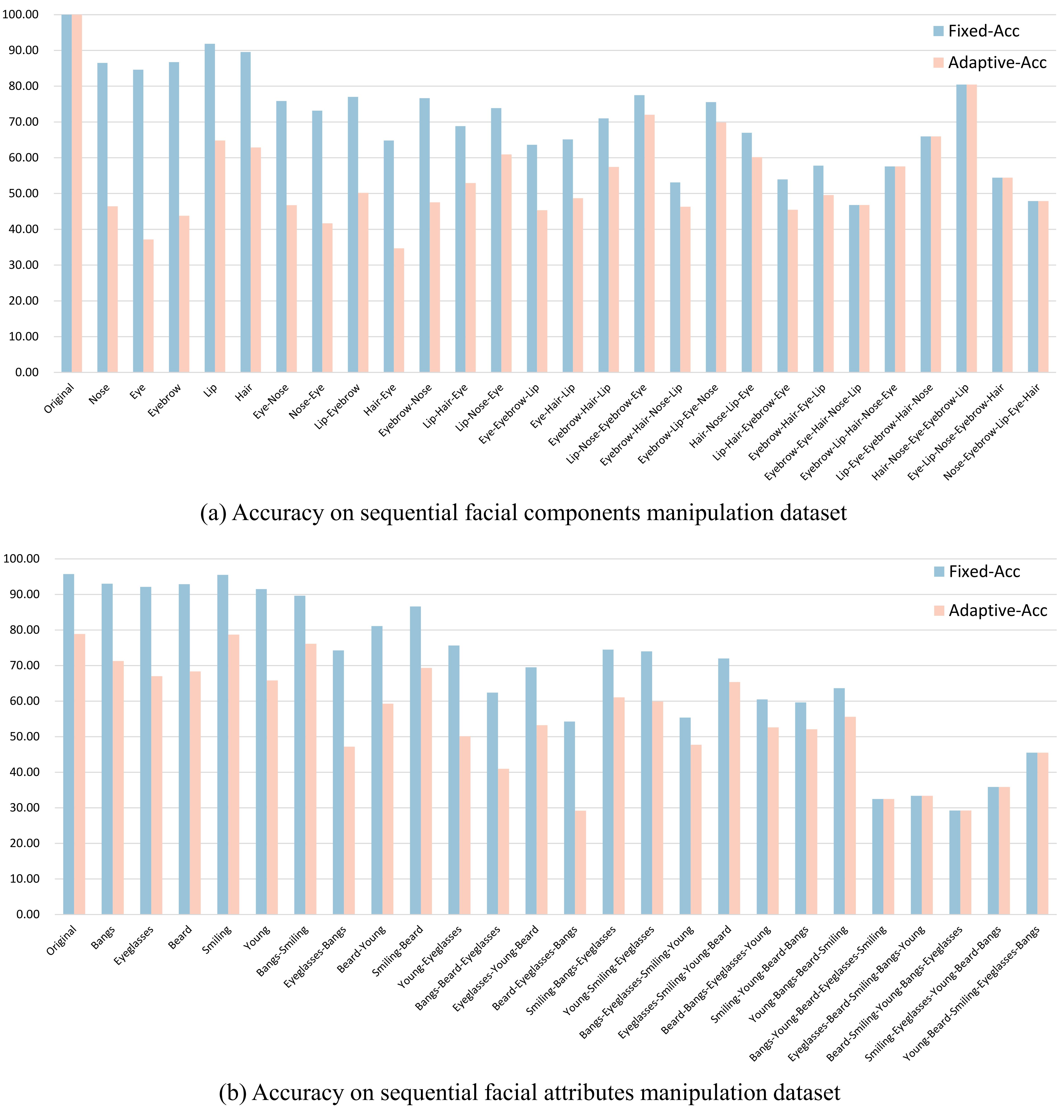}
	\end{center}
	\caption{\textcolor{black}{Accuracy for each manipulation sequence. Detection of a large portion of manipulation sequences is challenging and some extremely hard cases exist.}}
	\label{fig:acc}
\end{figure}

\begin{figure}[t] 
	\begin{center}
		\includegraphics[width=1\linewidth]{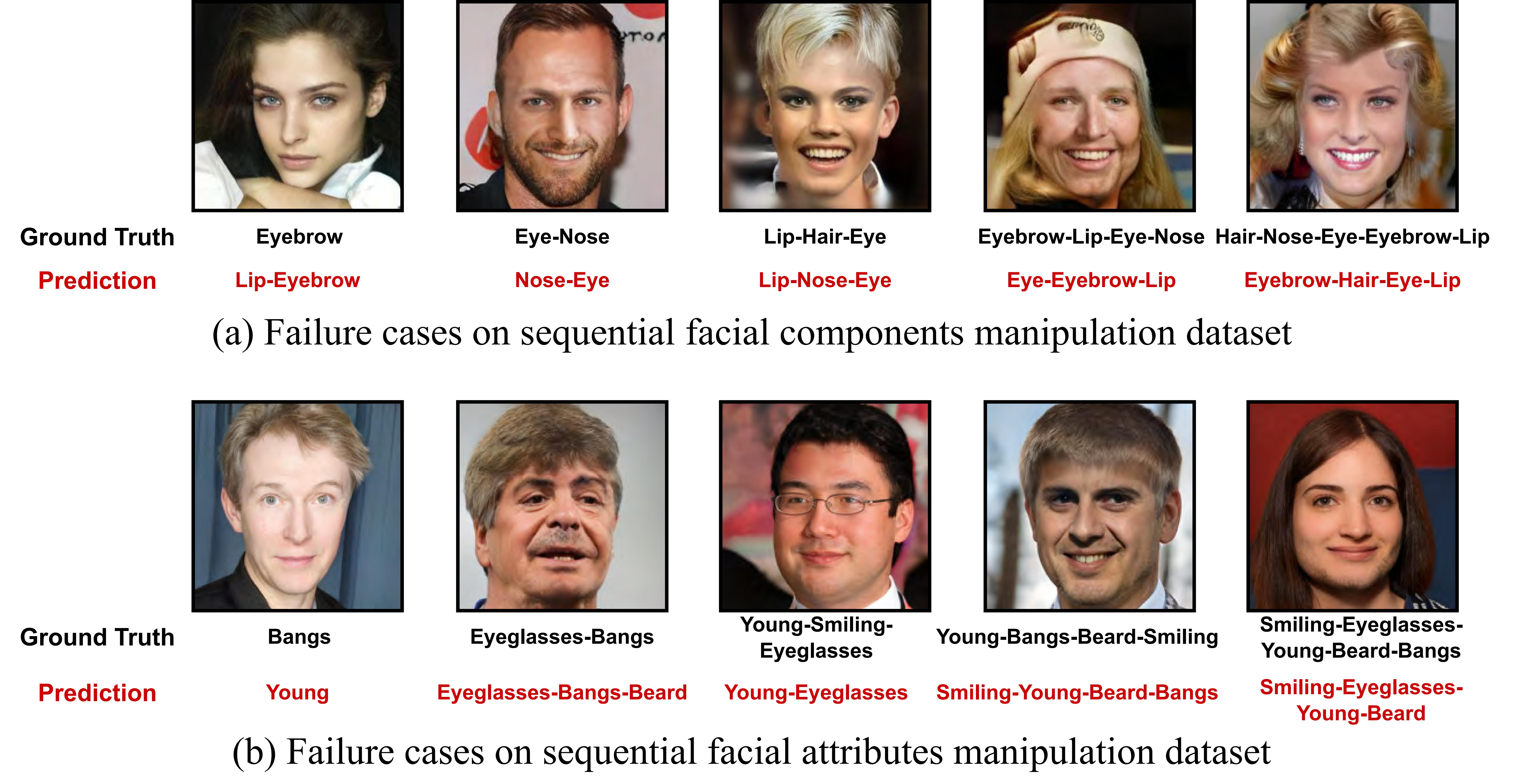}
	\end{center}
	\caption{\textcolor{black}{Examples of failure cases. Successful Seq-DeepFake detection is non-trivial as it demands the correct prediction in terms of manipulation types, orders and lengths simultaneously from hyper-realistic face images with subtle manipulation traces.}}
	\label{fig:failure}
    \vspace{-12pt}
\end{figure}

\begin{figure*}[t]
    \centering
    \includegraphics[width=1\linewidth]{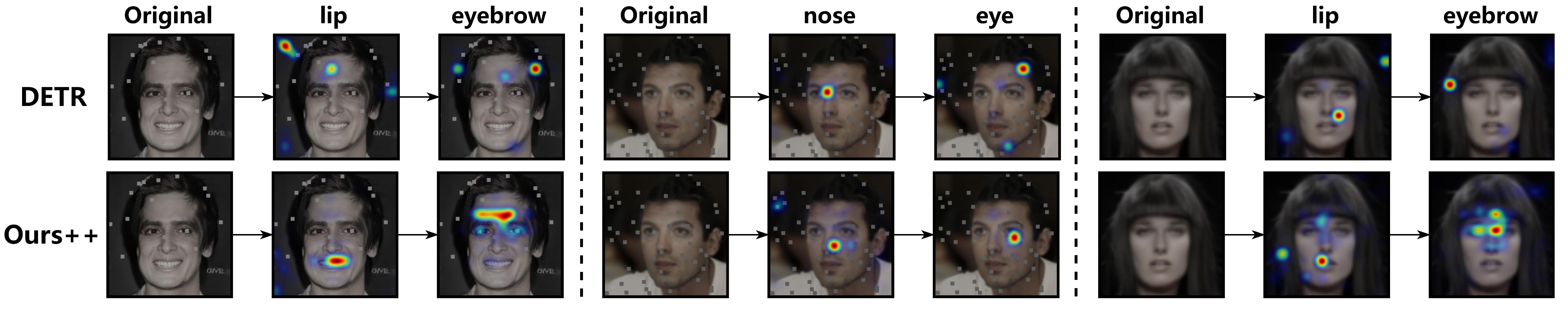}
    \caption{\textcolor{black}{Grad-CAM visualizations on the cross-attention maps corresponding to manipulation operations.}}
    \label{fig:gradcam}
\end{figure*}

\begin{figure*}[t]
    \centering
    \begin{minipage}[t]{0.75\textwidth}
    \centering
    \includegraphics[width=1\linewidth]{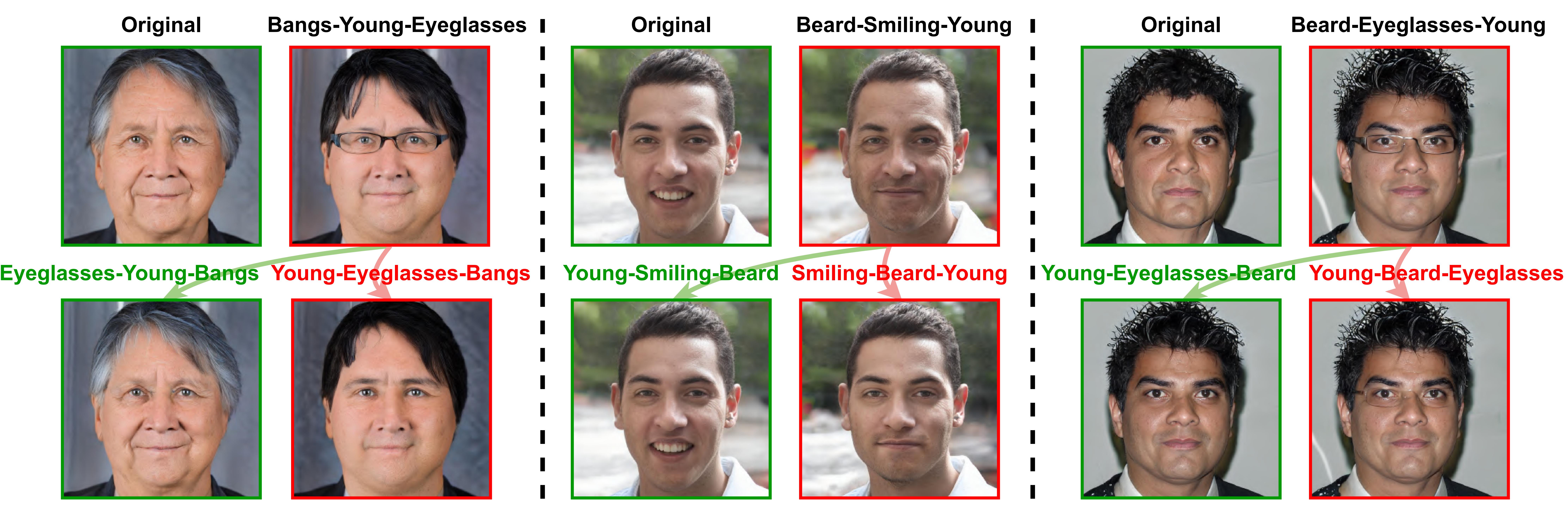}
    \caption{Face recovery based on correct and wrong facial manipulation sequences.}
    \label{fig:facerecover}
    \end{minipage}
    \begin{minipage}[t]{0.2\textwidth}
    \centering
    \includegraphics[width=1\linewidth]{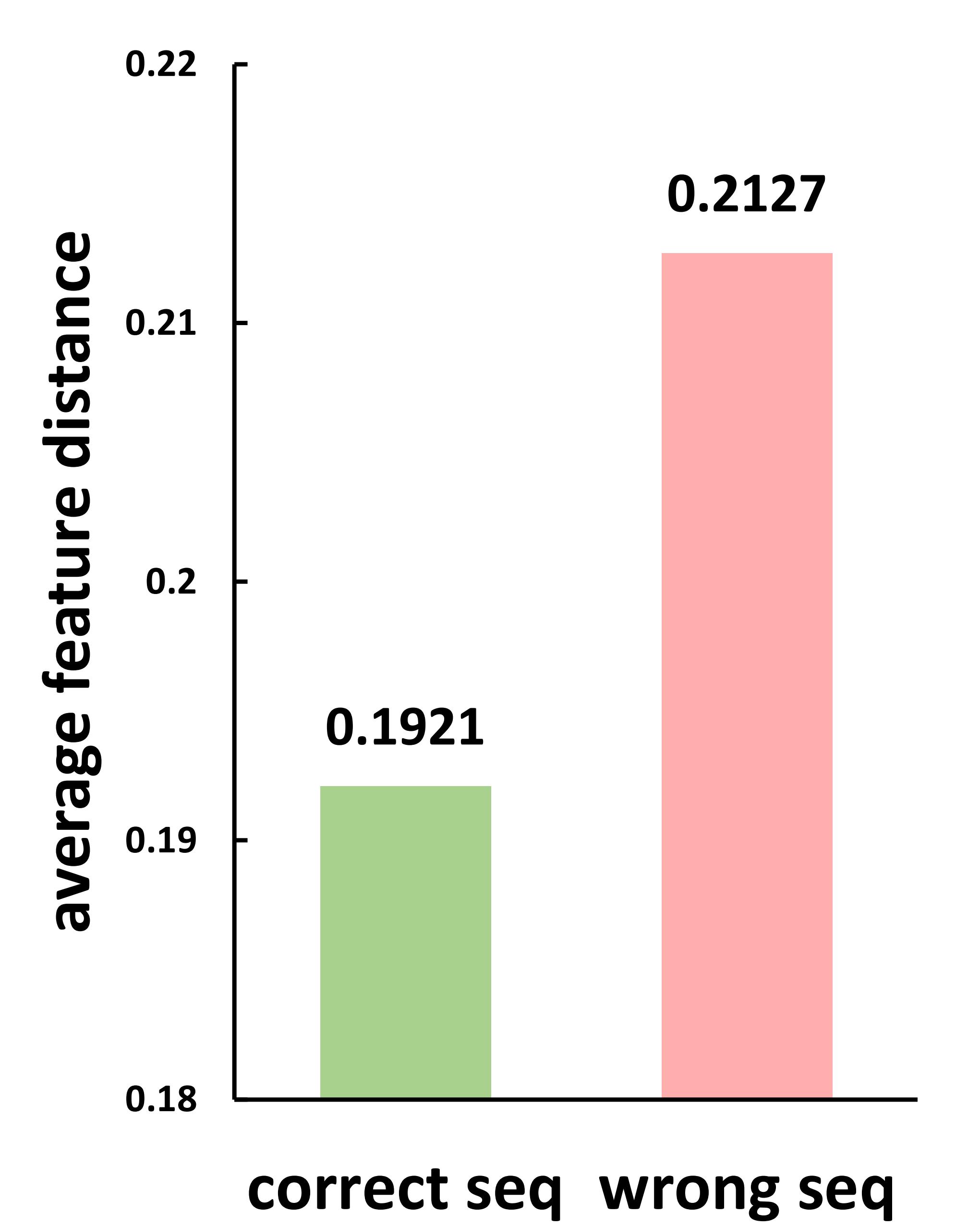}
    \caption{Identity preservation.}
    \label{fig:IDSocre}
    \end{minipage}
    \vspace{-10pt}
\end{figure*}

\textcolor{black}{As mentioned in Sec.~\ref{sec:seq-deepfake-dataset}, we generate 28 types of manipulation sequences based on facial components manipulation while 26 types of manipulation sequences based on facial attributes manipulation. To provide a more detailed analysis, in this section, we plot accuracy for each manipulation sequence in both facial manipulation methods as shown in Fig.~\ref{fig:acc}. It can be observed that diverse accuracy performance are achieved for different manipulation sequences, ranging from 46.81\% to 100\% under Fixed-Acc and 34.69\% to 100\% under Adaptive-Acc in sequential facial components manipulation, while ranging from 29.25\% to 95.75\% under Fixed-Acc and 29.21\% to 78.88\% under Adaptive-Acc in sequential facial attributes manipulation. This indicates a large portion of manipulation sequences are difficult to be detected and there exist some extremely hard cases. Therefore, we should further improve our method to cope with all types of manipulation sequences in the future. Furthermore, it can be seen from Fig.~\ref{fig:acc} that the accuracy gap between two evaluation metrics, Fixed-Acc and Adaptive-Acc, decreases along with the length of sequence increases. This is because the padded `no manipulation' class is fewer in the longer manipulation sequence when evaluating under Adaptive-Acc, which is closer to the evaluation under Fixed-Acc.}

\textcolor{black}{\subsection{Visualization of Attention Maps}
\label{sec:gradcam}
To demonstrate SeqFakeFormer++ detects the manipulation operations based on their sequence semantics, following ALBEF~\cite{li2021align}, we compute and visualize Grad-CAM~\cite{selvaraju2017grad} on the cross-attention maps in the first layer of Sequence Decoder for each manipulation operation in the sequence. We show some samples in Fig.~\ref{fig:gradcam} as follows. As displayed in this figure, SeqFakeFormer++ is able to detect manipulation operations by attending to semantically relevant facial regions, such as lip, nose, and eyebrow. In contrast, DETR randomly pays attention to some semantically irrelevant facial regions during the detection process. This indicates that SeqFakeFormer++ is more able to capture sequence semantics when detecting each manipulation operation, suggesting the feasibility of this application.}

\subsection{Face Recovery}
\label{sec:facerecovery}
After detecting facial manipulation sequences, we are able to perform more challenging tasks, like recovering the original face from the manipulated face image. Specifically, we formulate the Face Recovery task as: given a sequentially manipulated face image, reverse the manipulation process to get an image as close as possible to the original image. For example, in the facial attributes manipulation case, given an image generated by sequential manipulations on different attributes on the original face, we want to recover the original image. In fact, this task can be seen as an inverse sequential facial attribute manipulation problem, which can be effectively solved by the data generation pipeline described in Sec.~\ref{sec:seq-deepfake-dataset} in an inverse manner. Specifically, as can be observed in Fig.~\ref{fig:facerecover}, once we detect the correct facial manipulation sequence, \textit{i.e.,} correct manipulations ordered with correct manipulation steps, we can recover the original face by performing face attribute manipulation based on the inverse order of the detected sequence (process with green arrow). Comparatively, recovering the face image with wrongly ordered manipulation sequences may encounter different problems, such as incomplete recovery of age, smile, glasses, and \textit{etc}. (process with red arrow). Fig.~\ref{fig:IDSocre} evaluates the results using identity preservation metrics as in~\cite{jiang2021talk}, where smaller feature distance means identity is better preserved. The average feature distance between randomly selected 100 original faces and recovered faces using correct manipulation sequences is clearly smaller than that of the wrongly ordered sequence, implying the identity can be better recovered with correct manipulation sequence. Based on the above analysis and experiments, we prove that the detection of facial manipulation sequences is highly useful for face recovery, and we hope it can be applied to more meaningful tasks in the future.

\textcolor{black}{\subsection{Failure Cases}}

\textcolor{black}{To provide a deeper understanding for our novel task and method, we display some failure cases produced by the proposed method as illustrated in Fig.~\ref{fig:failure}. From Fig.~\ref{fig:failure}, it can be seen that there exist diverse failure cases, including wrong predictions with respect to manipulation type, sequence order, sequence length, and \textit{etc}. This validates that it is quite challenging for our novel research problem as we need to detect facial manipulation sequences in terms of correct manipulation types, orders and lengths simultaneously from hyper-realistic face images with subtle manipulation traces. This motivates us to continually develop detection methods to tackle such a novel yet challenging research problem.}

\begin{table}[t]
\scriptsize
\renewcommand{\arraystretch}{1}
\centering
\caption{\textcolor{black}{Accuracy of detecting Seq-DeepFake-P based on sequential facial components manipulation with perturbations.}}
\textcolor{black}{
\resizebox{\linewidth}{!}{%
\begin{tabular}{cl|cc|cc|cc}
\Xhline{2.5\arrayrulewidth}
\multicolumn{2}{c|}{\multirow{2}{*}{Methods}} & \multicolumn{2}{c|}{\textcolor{black}{ResNet-18}}  & \multicolumn{2}{c|}{ResNet-34}  & \multicolumn{2}{c}{RestNet-50}  \\ \cline{3-8} 
\multicolumn{2}{c|}{}          & \textcolor{black}{Fixed-Acc} & \textcolor{black}{Adap-Acc} & Fixed-Acc & Adap-Acc & Fixed-Acc & Adap-Acc \\ \hline
\multicolumn{2}{c|}{Multi-Cls} & \textcolor{black}{60.90}     & \textcolor{black}{40.44}    & 63.45     & 40.90    & 64.58     & 42.21    \\
\multicolumn{2}{c|}{DETR~\cite{carion2020end}}      & \textcolor{black}{60.86}     & \textcolor{black}{38.81}    & 63.13     & 40.78    & 61.17     & 38.19    \\
\multicolumn{2}{c|}{Ours~\cite{shao2022detecting}}      & \textcolor{black}{60.97}     & \textcolor{black}{40.90}    & 63.51     & 42.01    & 65.19     & 44.34    \\
\multicolumn{2}{c|}{Ours++}                   & \textcolor{black}{\textbf{63.28}} & \textcolor{black}{\textbf{42.56}} & \textbf{65.23} & \textbf{44.45} & \textbf{66.11} & \textbf{45.35} \\
\Xhline{2.5\arrayrulewidth}
\end{tabular}}
}
\label{tbl:facecomp++}
\end{table}

\begin{table}[t]
\scriptsize
\renewcommand{\arraystretch}{1}
\centering
\caption{\textcolor{black}{Accuracy of detecting Seq-DeepFake-P based on sequential facial attributes manipulation with perturbations.}}
\textcolor{black}{
\resizebox{\linewidth}{!}{%
\begin{tabular}{cl|cc|cc|cc}
\Xhline{2.5\arrayrulewidth}
\multicolumn{2}{c|}{\multirow{2}{*}{Methods}} & \multicolumn{2}{c|}{\textcolor{black}{ResNet-18}}  & \multicolumn{2}{c|}{ResNet-34}  & \multicolumn{2}{c}{RestNet-50}  \\ \cline{3-8} 
\multicolumn{2}{c|}{}          & \textcolor{black}{Fixed-Acc} & \textcolor{black}{Adap-Acc} & Fixed-Acc & Adap-Acc & Fixed-Acc & Adap-Acc \\ \hline
\multicolumn{2}{c|}{Multi-Cls} & \textcolor{black}{62.95}     & \textcolor{black}{41.16}    & 64.60     & 43.14    & 64.77     & 43.47    \\
\multicolumn{2}{c|}{DETR~\cite{carion2020end}}      & \textcolor{black}{63.08}     & \textcolor{black}{40.69}    & 65.65     & 44.93    & 65.62     & 45.23    \\
\multicolumn{2}{c|}{Ours~\cite{shao2022detecting}}      & \textcolor{black}{63.12}     & \textcolor{black}{43.56}    & 65.57     & 44.86    & 65.62     & 45.43    \\
\multicolumn{2}{c|}{Ours++}                   & \textcolor{black}{\textbf{65.60}} & \textcolor{black}{\textbf{45.11}} & \textbf{66.56} & \textbf{46.42} & \textbf{66.64} & \textbf{47.30} \\
\Xhline{2.5\arrayrulewidth}
\end{tabular}}}
\label{tbl:faceattr++}
\vspace{-12pt}
\end{table}

\begin{table}[t]
\scriptsize
\renewcommand{\arraystretch}{1}
\centering
\caption{\textcolor{black}{Accuracy of detecting Seq-DeepFake-P compared to deepfake detection methods.}}
\textcolor{black}{\begin{tabular}{c|cc|cc}
\Xhline{2.5\arrayrulewidth}
\multirow{2}{*}{Methods} & \multicolumn{2}{c|}{Face Components Manipulation}    & \multicolumn{2}{c}{Face Attributes Manipulation}     \\ \cline{2-5} 
                         & \multicolumn{1}{c|}{Fixed-Acc}      & Adap-Acc   & \multicolumn{1}{c|}{Fixed-Acc}      & Adap-Acc   \\ \hline
DRN~\cite{wang2019detecting}               & \multicolumn{1}{c|}{56.23}          & 30.07         & \multicolumn{1}{c|}{61.23}          & 38.33          \\
MA~\cite{zhao2021multi}                     & \multicolumn{1}{c|}{65.41}          & 43.88          & \multicolumn{1}{c|}{65.59}          & 44.66      \\
TS~\cite{luo2021generalizing}               & \multicolumn{1}{c|}{64.76}          & 42.89          & \multicolumn{1}{c|}{65.56}          & 44.72          \\
Ours~\cite{shao2022detecting}                     & \multicolumn{1}{c|}{65.19} & 44.34 & \multicolumn{1}{c|}{65.62} & 45.43 \\ 
Ours++                     & \multicolumn{1}{c|}{\textbf{66.11}} & \textbf{45.35} & \multicolumn{1}{c|}{\textbf{66.64}} & \textbf{47.30} \\\Xhline{2.5\arrayrulewidth}
\end{tabular}}
\label{tbl:deepfakebaseline++}
\end{table}

\begin{table}[t]
\scriptsize
\renewcommand{\arraystretch}{1}
\centering
\caption{\textcolor{black}{Accuracy of detecting Seq-DeepFake-P compared to multimodal learning methods.}}
\textcolor{black}{\begin{tabular}{c|cc|cc}
\Xhline{2.5\arrayrulewidth}
\multirow{2}{*}{Methods} & \multicolumn{2}{c|}{Face Components Manipulation}    & \multicolumn{2}{c}{Face Attributes Manipulation}     \\ \cline{2-5} 
                         & \multicolumn{1}{c|}{Fixed-Acc}      & Adap-Acc   & \multicolumn{1}{c|}{Fixed-Acc}      & Adap-Acc  \\ \hline
\textcolor{black}{BLIP~\cite{li2022blip}   }            & \multicolumn{1}{c|}{\textcolor{black}{55.19}}          & \textcolor{black}{37.50}          & \multicolumn{1}{c|}{\textcolor{black}{57.78}}          & \textcolor{black}{39.97}          \\
ALBEF~\cite{li2021align}                     & \multicolumn{1}{c|}{60.78} & 38.15 & \multicolumn{1}{c|}{60.23} & 42.02 \\ 
Ours~\cite{shao2022detecting}                     & \multicolumn{1}{c|}{65.19} & 44.34 & \multicolumn{1}{c|}{65.62} & 45.43 \\ 
Ours++                     & \multicolumn{1}{c|}{\textbf{66.11}} & \textbf{45.35} & \multicolumn{1}{c|}{\textbf{66.64}} & \textbf{47.30} \\\Xhline{2.5\arrayrulewidth}
\end{tabular}}
\label{tbl:multimodal_compare}
\end{table}

\begin{table}[t]
\scriptsize
\renewcommand{\arraystretch}{1}
\centering
\caption{\textcolor{black}{Ablation study of SeqFakeFormer++ in detecting Seq-DeepFake-P based on sequential facial components manipulation.}}
\begin{tabular}{cc|cc|cc}
\Xhline{2.5\arrayrulewidth}
\multicolumn{2}{c|}{Components}                      & \multicolumn{2}{c|}{ResNet-34}                       & \multicolumn{2}{c}{ResNet-50}                        \\ \hline
\multicolumn{1}{c|}{ISM} & ISC          & \multicolumn{1}{c|}{Fixed-Acc}      & Adap-Acc   & \multicolumn{1}{c|}{Fixed-Acc}      & Adap-Acc   \\ \hline
\multicolumn{1}{c|}{\XSolidBrush}     & \XSolidBrush   & \multicolumn{1}{c|}{63.51}          & 42.01         & \multicolumn{1}{c|}{65.19}          & 44.34             \\
\multicolumn{1}{c|}{\XSolidBrush}     & \CheckmarkBold & \multicolumn{1}{c|}{63.72} & 42.13 & \multicolumn{1}{c|}{65.22} & 44.50          \\
\multicolumn{1}{c|}{\CheckmarkBold}   & \XSolidBrush   & \multicolumn{1}{c|}{63.66} & 42.27 & \multicolumn{1}{c|}{65.34} & 44.52          \\
\multicolumn{1}{c|}{\CheckmarkBold}   & \CheckmarkBold & \multicolumn{1}{c|}{\textbf{65.23}} & \textbf{44.45} & \multicolumn{1}{c|}{\textbf{66.11}} & \textbf{45.35} \\ \Xhline{2.5\arrayrulewidth}
\end{tabular}
\label{tbl:Ablfacecomp++}
\end{table}

\begin{table}[t]
\scriptsize
\renewcommand{\arraystretch}{1}
\centering
\caption{\textcolor{black}{Ablation study of SeqFakeFormer++ in detecting Seq-DeepFake-P based on sequential facial attributes manipulation.}}
\begin{tabular}{cc|cc|cc}
\Xhline{2.5\arrayrulewidth}
\multicolumn{2}{c|}{Components}                      & \multicolumn{2}{c|}{ResNet-34}                       & \multicolumn{2}{c}{ResNet-50}                        \\ \hline
\multicolumn{1}{c|}{ISM} & ISC          & \multicolumn{1}{c|}{Fixed-Acc}      & Adap-Acc   & \multicolumn{1}{c|}{Fixed-Acc}      & Adap-Acc   \\ \hline
\multicolumn{1}{c|}{\XSolidBrush}     & \XSolidBrush   & \multicolumn{1}{c|}{65.57}          & 44.86         & \multicolumn{1}{c|}{65.62}          & 45.43          \\
\multicolumn{1}{c|}{\XSolidBrush}     & \CheckmarkBold & \multicolumn{1}{c|}{66.17} & 46.36 & \multicolumn{1}{c|}{65.87} & 46.21          \\
\multicolumn{1}{c|}{\CheckmarkBold}   & \XSolidBrush   & \multicolumn{1}{c|}{66.47} & 46.31 & \multicolumn{1}{c|}{66.48} & 46.24          \\
\multicolumn{1}{c|}{\CheckmarkBold}   & \CheckmarkBold & \multicolumn{1}{c|}{\textbf{66.56}} & \textbf{46.42} & \multicolumn{1}{c|}{\textbf{66.64}} & \textbf{47.30} \\ \Xhline{2.5\arrayrulewidth}
\end{tabular}
\label{tbl:Ablfaceattr++}
\vspace{-10pt}
\end{table}

\textcolor{black}{\section{Benchmark for Seq-DeepFake-P}}
\textcolor{black}{We tabulate the benchmark for detecting sequential facial manipulation with perturbations (Seq-DeepFake-P) based on facial components manipulation and facial attributes manipulation in Tables~\ref{tbl:facecomp++} to \ref{tbl:deepfakebaseline++}. Compared to benchmark results regarding Seq-DeepFake dataset in Tables~\ref{tbl:facecomp} to \ref{tbl:deepfakebaseline}, Tables~\ref{tbl:facecomp++} to \ref{tbl:deepfakebaseline++} show that the performance obtained by all the baselines and proposed method degrades when dealing with Seq-DeepFake-P, especially in the detection of sequential facial components manipulation. This demonstrates it is much more challenging to detect Seq-DeepFake in the more realistic scenario under various perturbations.}

Facing such a harder scenario, it can be seen from Tables~\ref{tbl:facecomp++} and \ref{tbl:faceattr++} that the proposed SeqFakeFormer++ (Ours++) obtains the best 
detection performance for both facial components manipulation and facial attributes manipulation. It surpasses all considered baselines by 1-4\% under two evaluation metrics. Particularly, we can clearly observe that SeqFakeFormer++ substantially outperforms SeqFakeFormer (Ours) in Tables~\ref{tbl:facecomp++} and \ref{tbl:faceattr++}. This implies the efficacy of two supplemented image-sequence reasoning modules for tackling Seq-DeepFake-P data. \textcolor{black}{In addition, we further incorporate one more image feature extraction backbone, ResNet-18, into the proposed framework, as shown in Tables~\ref{tbl:facecomp++} and \ref{tbl:faceattr++}. It can be seen that SeqFakeFormer and SeqFakeFormer++ with ResNet-18~\cite{he2016deep} backbone also yield better performance compared to other baselines, presenting similar trends as the other backbones. This further indicates the versatility of the proposed method that SeqFakeFormer and SeqFakeFormer++ are compatible with various image feature extraction backbones.}

\textcolor{black}{Meanwhile, we tabulate the comparison with three SOTA deepfake detection methods, including our conference version, in Table~\ref{tbl:deepfakebaseline++}. SeqFakeFormer++ also exceeds all SOTA deepfake detection methods in both sequential manipulation detection tasks. Notably, SeqFakeFormer++ is able to improve the performance with respect to Adaptive-Acc by at least 3\% compared to these three SOTA deepfake detection methods.}

\textcolor{black}{To further validate the superiority of the proposed method over some relevant multimodal learning methods, we fine-tune representative multimodal models such as BLIP~\cite{li2022blip}  and ALBEF~\cite{li2021align} with Seq-DeepFake-P data and tabulate their comparison results in Table~\ref{tbl:multimodal_compare}. The experimental results demonstrate that SeqFakeFormer and SeqFakeFormer++ surpass both multimodal baselines by a large margin for the task of Seq-DeepFake, suggesting the efficacy of the proposed method.}


\textcolor{black}{\subsection{Ablation Study}}
\textcolor{black}{In this sub-section we study the impact of two proposed image-sequence reasoning modules in SeqFakeFormer++, Image-Sequence Contrastive Learning (ISC) and Image-Sequence Matching (ISM), to the overall performance. We tabulate ablation studies in terms of ISM and ISC in Tables~\ref{tbl:Ablfacecomp++} and \ref{tbl:Ablfaceattr++}. As observed from Tables~\ref{tbl:Ablfacecomp++} and \ref{tbl:Ablfaceattr++}, the performance regarding two Seq-DeepFake detection tasks under all evaluation metrics degenerates once removing one of these reasoning modules, indicating the effectiveness and complementarity of them. It should be noted from  Tables~\ref{tbl:Ablfacecomp++} and \ref{tbl:Ablfaceattr++} that adding one of the reasoning modules in SeqFakeFormer++ will obtain better performance than SeqFakeFormer. This further verifies the importance of each reasoning module for detecting Seq-DeepFake under perturbations.}

\section{Conclusion}
This paper studies a novel research problem -- Detecting Sequential DeepFake Manipulation, aiming to detect a sequential vector of multi-step facial manipulation operations. We also introduce the first Seq-DeepFake dataset to provide sequentially manipulated face images. Supported by this new dataset, we cast detecting Seq-DeepFake manipulation as a specific image-to-sequence task and propose a Seq-DeepFake Transformer (SeqFakeFormer). Two modules, Spatial Relation Extraction and Sequential Relation Modeling with Spatially Enhanced Cross-Attention, are integrated into SeqFakeFormer, complementing each other. \textcolor{black}{We further construct a more challenging Seq-DeepFake-P dataset to mimic the more realistic deepfake data distributions under perturbations. A dedicated Seq-DeepFake Transformer with Image-Sequence Reasoning (SeqFakeFormer++) is built to deal with such a noisy scenario. Extensive experimental results demonstrate the superiority of SeqFakeFormer and SeqFakeFormer++ and valuable observations pave the way for future research in broader deepfake detection.}

\section{acknowledgements}
This study is supported by National Natural Science Foundation of China (Grant No. 62306090); Natural Science Foundation of Guangdong Province of China (Grant No. 2024A1515010147); This study is supported by the Ministry of Education, Singapore, under its MOE AcRF Tier 2 (MOET2EP20221-0012), NTU NAP, and under the RIE2020 Industry Alignment Fund – Industry Collaboration Projects (IAF-ICP) Funding Initiative, as well as cash and in-kind contribution from the industry partner(s).

\section{Data Availability Statement}
The Seq-DeepFake dataset analysed during this study is publicly available for the research purpose - \href{https://github.com/rshaojimmy/SeqDeepFake}{Seq-DeepFake dataset}.


%
%

\bibliographystyle{spmpsci}      
\bibliography{ref}   


\end{document}